\documentclass{article}

\PassOptionsToPackage{numbers}{natbib}

\usepackage[final]{neurips_2024}

\usepackage{float}

\usepackage[utf8]{inputenc} 
\usepackage[T1]{fontenc}    
\usepackage{hyperref}       
\usepackage{url}            
\usepackage{booktabs}       
\usepackage{amsfonts}       
\usepackage{nicefrac}       
\usepackage{microtype}      
\usepackage{xcolor}         
\usepackage{multirow}
\usepackage{xcolor,colortbl}
\usepackage{adjustbox}
\usepackage{siunitx}
\usepackage[font=small]{caption}
\usepackage{placeins}
\usepackage{makecell}

\title{Atlas 2 - Foundation models for clinical deployment}

\author{
Maximilian Alber$^{\;*\;\ddag\;1\;13}$,
Timo Milbich$^{\;*\;1}$,
Alexandra Carpen-Amarie$^{\;*\;1}$,\AND
Stephan Tietz$^{\;\#\;1}$, 
Jonas Dippel$^{\;\#\;1\;7\;8}$,
Lukas Muttenthaler$^{\;\#\;1\;15\;16}$, \AND
Beatriz Perez Cancer$^{\;\#\;1}$, 
Alessandro Benetti$^{\;\#\;1}$, 
Panos Korfiatis$^{\;\#\;3}$, \AND
Elias Eulig$^{\;1}$, 
Jérôme Lüscher$^{\;1}$, 
Jiasen Wu$^{\;1}$, 
Sayed Abid Hashimi$^{\;1}$, \AND
Gabriel Dernbach$^{\;1\;8\;13}$,
Simon Schallenberg$^{\;13}$,
Neelay Shah$^{\;1}$, \AND
Moritz Krügener$^{\;1}$,
Aniruddh Jammoria$^{\;1}$,
Jake Matras$^{\;5}$, 
Patrick Duffy$^{\;6}$,\AND
Matt Redlon$^{\;4}$,
Philipp Jurmeister$^{\;11\;12}$,
David Horst$^{\;11\;13}$,
Lukas Ruff$^{\;1}$,\AND
Klaus-Robert Müller$^{\dag\;7\;8\;9\;10}$,
Frederick Klauschen$^{\dag\;8\;11\;12\;13\;14}$,
Andrew Norgan$^{\dag\;2}$\\ \\
$^1$ Aignostics, Germany \AND 
$^2$ Department of Laboratory Medicine and Pathology, Mayo Clinic, Rochester, MN, US \AND
    $^3$ Department of Radiology,  Mayo Clinic, Rochester MN, US \AND 
    $^4$ Department of Information Technology, Mayo Clinic, Rochester MN, US \AND 
    $^5$ Mayo Clinic, Rochester MN, US \AND 
    $^6$ Digital Pathology, Mayo Clinic, Rochester MN, US \AND 
    $^7$ Machine Learning Group, Technische Universität Berlin, Germany \AND 
    $^8$ BIFOLD – Berlin Institute for the Foundations of Learning and Data, Germany \AND 
    $^9$ Department of Artificial Intelligence, Korea University, Republic of Korea \AND 
    $^{10}$ Max-Planck Institute for Informatics, Germany \AND 
    $^{11}$ German Cancer Research Center (DKFZ) \& German Cancer Consortium (DKTK), \\ Berlin \& Munich Partner Sites, Germany \AND 
    $^{12}$ Institute of Pathology, Ludwig-Maximilians-Universität München, Germany \AND 
    $^{13}$ Institute of Pathology, Charité – Universitätsmedizin Berlin, Germany \AND 
    $^{14}$ Bavarian Cancer Research Center (BZKF), Germany \AND
    $^{15}$ Helmholtz Munich, Germany \AND
    $^{16}$ Technical University Munich, Germany \AND
    $^{*, \#, \dag}$ Equal contribution respectively \AND
    $\ddag$ Corresponding author}

\begin{document}

\maketitle

\begin{abstract}
Pathology foundation models substantially advanced the possibilities in computational pathology --- yet tradeoffs in terms of performance, robustness, and computational requirements remained, which limited their clinical deployment.
In this report, we present Atlas 2, Atlas 2-B, and Atlas 2-S, three pathology vision foundation models which bridge these shortcomings by showing state-of-the-art prediction performance, robustness, and resource efficiency in a comprehensive evaluation across eighty public benchmarks. Our models were trained on the largest pathology foundation model dataset to date comprising 5.5 million histopathology whole slide images, collected from three medical institutions Charité - Universitätsmedizin Berlin, LMU Munich, and Mayo Clinic.
\end{abstract}

\begin{figure}[h!]
\centering
        \includegraphics[width=\textwidth]{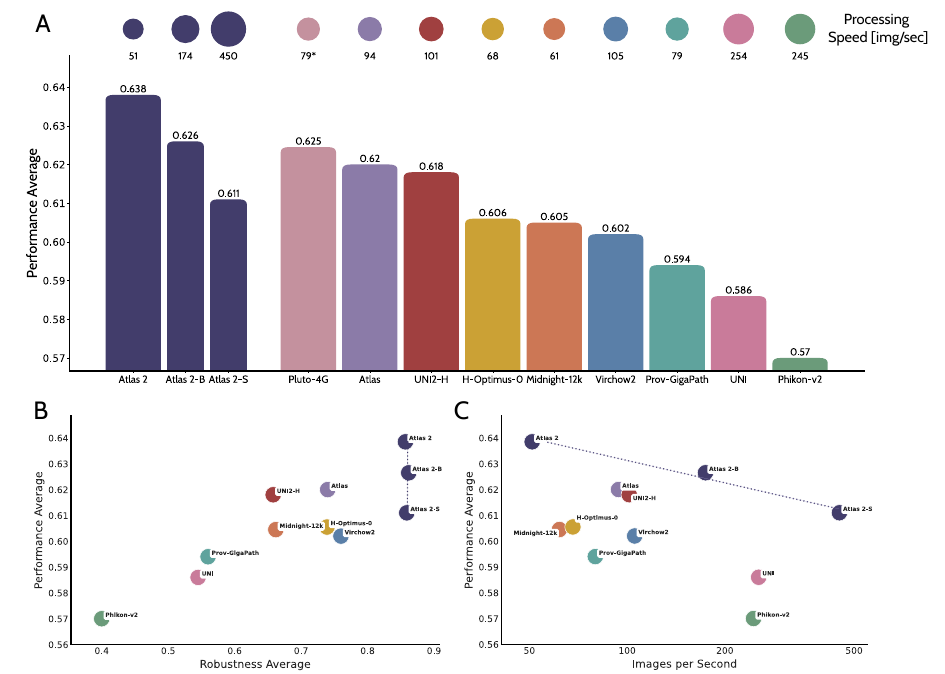}
    \caption{\textbf{(A)}  The results show that Atlas 2 is the best-performing model. Additionally Atlas 2-B and 2-S have similar prediction performance as contenders but are up to a magnitude more efficient. The performance average is based on morphology prediction tasks from \textit{eva}~\cite{kaiko.ai2024eva} and molecular prediction tasks from HEST~\cite{jaume2024hest} in Table~\ref{tab:main_results}. The dots show the processing speed of the respective model for a 224$\times$224 pixel image on an L4 GPU. The value for Pluto-4G~\cite{pluto4g} is approximated by taking the fastest speed of same sized models~\cite{kaplan2025openmidnight, hoptimus0, xu_whole-slide_2024}. \textbf{(B)} The results show that Atlas 2, Atlas 2-B, and Atlas 2-S are the most robust models and also the state-of-the-art performance-robustness Pareto front. The plot shows the performance average from (A) and the robustness average from Table~\ref{tab:distillation_results}. \textbf{(C)} The data shows that the Atlas 2 models are the performance-resource efficiency Pareto front. The plot shows the performance average of (A).}
    \label{fig:overview}
\end{figure}

\section{Introduction}

\begin{figure}[t!]
\centering
        \includegraphics[width=\textwidth]{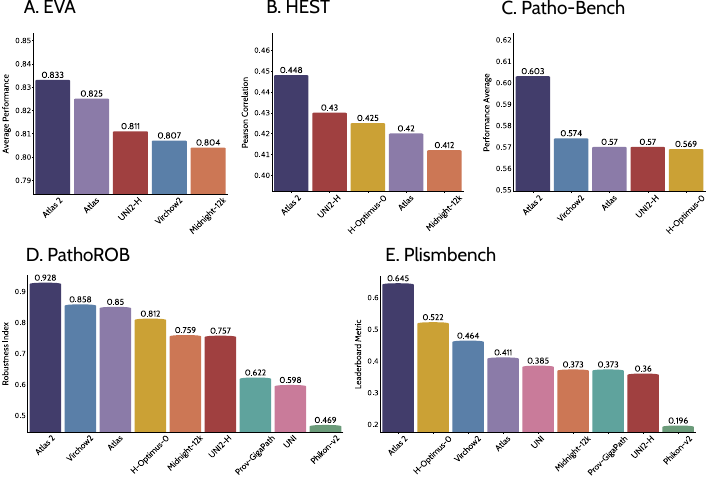}
    \caption{(A)--(E) show the average performance per evaluation framework~\cite{kaiko.ai2024eva}, \cite{jaume2024hest}, \cite{pathobench}, \cite{koemen2025pathorob}, and \cite{plismbench}. Atlas 2 is the leading contender on all frameworks.}
    \label{fig:all_frameworks}
\end{figure}

Digital and computational pathology including large-scale slide digitization, virtual microscopy for diagnostics, and artificial intelligence (AI)-based histological image analysis has started to transform anatomic pathology and led to promising proofs of concept and applications \citep[e.g.,][]{klauschen2024toward,diao2021human,raciti2023clinical,binder2021morphological,keyl2022patient}. However, generalization and robustness remain challenging and have hindered the broad translation of AI applications into clinical routine diagnostics. Despite increasing digitization efforts, scarcity of training data, particularly for infrequent and rare diseases~\cite{dippel2024ai}, remains a challenge. Furthermore, generating sufficient labeled data representing the full spectrum of human disease, biological, and technical variability inherent to morphology, tissue processing, staining, and scanners has proven logistically and financially challenging. Addressing these problems, pathology foundation models have recently gained traction based on their promise to achieve robust and generalizable data representations by incorporating the diversity present in pathology through large-scale self-supervised training. However, tradeoffs in terms of performance, robustness, and computational requirements remained and have so far limited their clinical impact. In this report, we present an updated version of our previously published pathology foundation model Atlas~\cite{alber2025atlas} that addresses these deficiencies.
To this end, we substantially expand the underlying database by using a multi-centric corpus of 5.5 million histopathology whole slide images derived from Charité - Universitätsmedizin Berlin, LMU Munich, and Mayo Clinic, increase the model parameter count, and improve our training recipe to develop a novel pathology foundation model ``Atlas 2''. We distill our main model into the lightweight versions ``Atlas 2-B'' and ``Atlas 2-S'', which are based on the more resource efficient architectures ViT-B and ViT-S. They can process a standard-size image 3.4 and 9 times faster, respectively. All models incorporate a broad diversity of diseases, staining types, and scanners, and utilize multiple image magnifications during training. We compare the performance of Atlas 2 to other leading models available for testing using eighty benchmarks assessing a variety of downstream pathology tasks. An overview of model characteristics and main results can be found in Figure~\ref{fig:overview} and Figure~\ref{fig:all_frameworks}.

\section{Related Work}

\paragraph{Pathology foundation models} In the past years pathology foundation models~\cite{dippel_rudolfv_2024,alber2025atlas,zimmermann_virchow_2024,filiot_phikon-v2_2024,xu_whole-slide_2024,hoptimus0,chen_towards_2024,ding2025multimodal,pluto4g,lenz2024unsupervisedfoundationmodelagnosticslidelevel,juyal_pluto_2024,lu2024avisionlanguage,Qiu_2025_ICCV} established themselves as the basis for developing clinical-grade applications~\cite{wang2024pathologychief} in digital pathology. These models are divided into tile- or slide-based models. 
Tile-based models~\cite{dippel_rudolfv_2024,alber2025atlas,zimmermann_virchow_2024,chen_towards_2024,hoptimus1,pluto4g,KDK_Training_MICCAI2025,kaplan2025openmidnight} are trained via self-supervised-learning frameworks such as DINOv2~\cite{oquab2023dinov2} or DINOv3~\cite{simeoni2025dinov3} to encode small, fixed-sized image tiles into compact embeddings. These embeddings then typically serve as a starting point to train slide-based pathology foundation models~\cite{lu2024avisionlanguage,shaikovski_prism_2024,wang2024pathologychief} which learn to aggregate the per-tile information into a single representation for an entire whole slide image (WSI) using slide-level supervision such as patient report-~\cite{lu2024avisionlanguage,vorontsov2025prism2unlockingmultimodalgeneral,shaikovski_prism_2024,sun2025cpath} or molecular~\cite{vaidya2025threads,xu2025multimodal} data.
Key drivers of the success story of pathology foundation models are increasing both the number of WSIs powering foundation model training~\cite{alber2025atlas,hoptimus1,zimmermann_virchow_2024} and the capacity of such models by means of their encoder size~\cite{zimmermann_virchow_2024,hoptimus1,kaplan2025openmidnight}. The presented model Atlas 2 is the largest tile-based pathology foundation model to date with a total of 5.5 million WSIs being used to train a 2 billion parameter Vision Transformer~\cite{dosovitskiy2020vit}.

\paragraph{Robustness of pathology foundation models} While pathology foundation models are trained on increasingly large and diverse pretraining datasets, their lack of robustness to scanner, lab processing artifacts, and staining variations was observed to be still problematic~\cite{koemen2025pathorob, plismbench,dejong_pathology_fm_robustness,koemen2024batcheffects,gustafsson2024evaluating,lin2025unveiling,carloni2025pathology,chai2025impact,moellers2026mindgapcontinuousmagnification}. This can lead to a substantial decrease in performance when prediction labels are correlated with such confounding information in the representations~\cite{clever-hans,unsupervised-clever-hans,koemen2025pathorob}. Multiple investigations have analyzed these shortcomings and proposed benchmarks to measure the robustness in this regard~\cite{plismbench, koemen2025pathorob}. So far, only limited work has been performed in improving the robustness of pathology foundation models. Examples are stain normalization~\cite{macenko, reinhard}, data augmentation~\cite{drexlin2025medi}, projecting out confounder information~\cite{nguyen2025fmmap, koemen2025pathorob}, distillation~\cite{plismbench}, or applying methods during downstream training (e.g. DANN~\cite{dann,koemen2025pathorob}).

\paragraph{Efficient pathology foundation models for clinical deployment} To provide efficient alternatives to large-scale frontier pathology foundation models, several groups~\cite{zimmermann_virchow_2024,pluto4g,filiot2025distillingfoundationmodelsrobust} have released lightweight counterparts by either training small encoders directly from scratch~\cite{pluto4g,kang2022benchmarking,nechaev_hibou_2024,kaiko2024towardslarge} or performing knowledge distillation~\cite{hinton2015distillingknowledgeneuralnetwork,beyer2022knowledgedistillationgoodteacher} from larger foundation models~\cite{filiot2025distillingfoundationmodelsrobust,zimmermann_virchow_2024}.

\section{Data and Methods}

\subsection{Dataset and Preprocessing}
We curate a dataset of 5.5 million de-identified WSIs, derived from the digital archives of Charité - Universitätsmedizin Berlin, LMU Munich, and Mayo Clinic, from which we extract tiles at multiple resolutions, namely 0.25, 0.5, 1.0, and 2.0 microns per pixel, corresponding to objective microscopic magnifications of $40\times$, $20\times$, $10\times$, and $5\times$, respectively.

\subsection{Model Framework and Compute Environment}
We perform model training in several phases. All of our model architectures are based on the Vision Transformer (ViT)~\cite{dosovitskiy2020vit}, using a patch-token size of 8. For our main model, Atlas 2, we use a 2 billion parameter variant. For the distilled versions, Atlas 2-B and Atlas 2-S, we use the model sizes ``base'' (ViT-B; 86 million parameters) and ``small'' (ViT-S; 22 million parameters), respectively. Our training framework uses an adapted RudolfV~\cite{dippel_rudolfv_2024} and Atlas~\cite{alber2025atlas} approach as well as novel developments. Parts of the framework are based on DINOv2~\cite{oquab2023dinov2} and selective improvements from DINOv3~\cite{simeoni2025dinov3}. For model training, we use NVIDIA B200 GPUs and NVIDIA A100 GPUs.

\subsection{Evaluation Protocols}
\label{ssec:evaluation_protocols}
For reproducibility and comparability, publicly available frameworks were used for the evaluation: HEST~\cite{jaume2024hest}, \textit{eva}~\cite{kaiko.ai2024eva}, PathoROB~\cite{koemen2025pathorob}, Plismbench~\cite{plismbench}, and Patho-Bench~\cite{pathobench}, covering tile-level as well as slide-level tasks. To further extend the pool of tasks, we additionally report a number of results for public benchmarks based on an in-house evaluation framework.
To evaluate the quality of the foundation model embeddings, we keep the model encoders frozen and solely apply task-specific prediction heads.
Unless stated otherwise, we report results based on a concatenation of the CLS token and the mean over patch tokens (``CLS+MEAN''). Results for the CLS token can be found in Appendix~\ref{ssec:additional_results}. Input images are normalized using the specified normalization statistics of the respective models. We do not apply normalization to the output tokens unless it is specifically done by the evaluation framework.

We evaluate 15 foundation models using their publicly available weights: UNI2-H~\cite{chen_towards_2024}, \mbox{H-Optimus-0~\cite{hoptimus0}},  
Midnight-12k~\cite{kaplan2025openmidnight}, Virchow2~\cite{zimmermann_virchow_2024}, Prov-GigaPath~\cite{xu_whole-slide_2024}, UNI~\cite{chen_towards_2024}, Phikon-v2~\cite{filiot_phikon-v2_2024}, the ViT-B/8, ViT-B/16, ViT-S/8, and ViT-S/16 from~\cite{kaiko2024towardslarge}, Hibou-B~\cite{nechaev2024hibou}, Phikon~\cite{Filiot2023ScalingSSLforHistoWithMIM}, and the ViT-S/8 and ViT-S/16 from~\cite{kang2022benchmarking}. For five additional models, Pluto-4G~\cite{pluto4g}, Pluto-4S-8~\cite{pluto4g}, Pluto-4S-16~\cite{pluto4g}, H-Optimus-1~\cite{hoptimus1}, and H0-mini~\cite{plismbench}, we include comparisons based on published results, as model weights are either not publicly released or not accessible to us for evaluation. 

\paragraph{HEST} The HEST-Benchmark~\cite{jaume2024hest} is composed of tasks for gene expression prediction and designed as multivariate regression. We use the default evaluation protocol and implementation~\cite{hestgithub}. The protocol leverages a Ridge Regression with Principal Component Analysis (PCA) to solve the multivariate regression on the extracted foundation model embeddings. The results are based on the Pearson correlation coefficient as defined in~\cite{jaume2024hest}. Dataset details can be found in Appendix~\ref{ssec:task_descriptions}.

\paragraph{\textit{eva}} The \textit{eva} framework~\cite{kaiko.ai2024eva} comprises tasks for clinical prediction across various cancer types and pathology applications. Again, we use the default evaluation protocol and implementation~\cite{evagithub}. For patch-level classification tasks, a linear classification head is trained on the CLS or CLS+MEAN tokens. For patch-level segmentation tasks, segmentation is performed on the extracted patch token embeddings. For slide-level classification tasks, \textit{eva} applies the default Attention-based Multiple Instance Learning (ABMIL)~\cite{ilse2018attention} protocol.
We use balanced accuracy as a performance metric for all patch-level and slide-level classification tasks. For patch-level segmentation tasks, Dice score without background is reported. Evaluation results are presented using the test split when provided, and the validation split when a test split is unavailable. Dataset details can be found in Appendix~\ref{ssec:task_descriptions}.

\paragraph{PathoROB} The PathoROB Robustness Index benchmark~\cite{koemen2025pathorob} is designed to evaluate model robustness with respect to non-biological, confounding features. We use their default evaluation protocol and implementation~\cite{pathorobgithub}. Our average values for Virchow2 and Prov-GigaPath differ slightly from official values due to using bicubic image resizing instead of bilinear. This does not affect the ranking. We provide dataset details in Appendix~\ref{ssec:task_descriptions}.

\paragraph{Plismbench} The Plismbench benchmark~\cite{plismbench} measures the representation consistency across different scanning and staining conditions. We use the default evaluation protocol and implementation~\cite{plismbenchgithub}. 
We report the leaderboard metric, defined as the average of four metrics: cosine similarity for all pairs, top-10 accuracy for cross-scanner pairs, top-10 accuracy for cross-staining pairs, and top-10 accuracy for cross-scanner and cross-staining pairs and use $n=8,139$ tiles which is the reference value reported in the repository. The official Plismbench leaderboard~\cite{plismbenchgithub} reports results for either the CLS+MEAN or the CLS token per model. Here, we report results for both the CLS+MEAN (in Table~\ref{tab:main_results}) and CLS tokens (in Table~\ref{tab:main_results_cls}).
Further details can be found in Appendix~\ref{ssec:task_descriptions}.

\paragraph{Patho-Bench} The Patho-Bench framework~\cite{pathobench} is composed of 95 benchmark tasks covering seven subcategories: morphological subtyping, tumor microenvironment (TME) characterization, tumor grading, molecular subtyping, mutation prediction, treatment response assessment, and survival prediction. We evaluate a subset of 53 tasks drawn from multiple datasets using the ABMIL evaluation protocol using the default implementation~\cite{pathobenchgithub}. The task selection is determined by implementation difficulties and dataset access. Performance is evaluated using task-appropriate metrics defined in the Patho-Bench protocol: area under the receiver operating characteristic curve (AUROC) for binary classification tasks, balanced accuracy for multi-class classification tasks, concordance index (C-index) for survival prediction tasks, and quadratic weighted Cohen’s kappa for tumor grading tasks. 
While Patho-Bench fixes the data splits, variance can be a result of the initialization and optimization of the ABMIL prediction head. To mitigate this, each task is performed three times and we report the mean of performance metrics across all data folds and random seeds. Evaluation results are presented using the provided data splits and task metadata on HuggingFace~\footnote{\url{https://huggingface.co/datasets/MahmoodLab/Patho-Bench}}. Further details are reported in Appendix~\ref{ssec:task_descriptions}.

\paragraph{Additional benchmarks} 
The public benchmarks MSI CRC~\cite{kather2019deep,kaczmarzyk2023champkit}, MSI STAD~\cite{kather2019deep,kaczmarzyk2023champkit}, and TCGA Uniform~\cite{komura_tcga_uniform} are not part of any public evaluation framework. For these benchmarks, we report an evaluation based on an internal framework. For each benchmark, we apply linear probing using logistic regression. Results are computed for 5 seeds over data split, shuffling, and initialization of the prediction head per foundation model and task. The mean across seeds is reported. Dataset splits are described in detail in Appendix~\ref{ssec:task_descriptions}.

\section{Results}

\begin{table}[!ht]
\centering
\begin{adjustbox}{width=1\textwidth}
\scriptsize
\begin{tabular}{l>{\raggedleft\arraybackslash}p{0.5cm}>{\raggedleft\arraybackslash}p{0.5cm}>{\raggedleft\arraybackslash}p{0.5cm}>{\raggedleft\arraybackslash}p{0.5cm}>{\raggedleft\arraybackslash}p{0.5cm}>{\raggedleft\arraybackslash}p{0.5cm}>{\raggedleft\arraybackslash}p{0.5cm}>{\raggedleft\arraybackslash}p{0.5cm}>{\raggedleft\arraybackslash}p{0.5cm}>{\raggedleft\arraybackslash}p{0.5cm}}%
&\rotatebox{50}{\makebox[0pt][l]{\hspace{-3mm}\strut \makecell[r]{Atlas 2}}}&\rotatebox{50}{\makebox[0pt][l]{\hspace{-3mm}\strut \makecell[r]{Pluto-4G*}}}&\rotatebox{50}{\makebox[0pt][l]{\hspace{-3mm}\strut \makecell[r]{Atlas}}}&\rotatebox{50}{\makebox[0pt][l]{\hspace{-3mm}\strut \makecell[r]{UNI2-H}}}&\rotatebox{50}{\makebox[0pt][l]{\hspace{-3mm}\strut \makecell[r]{H-Optimus-0}}}&\rotatebox{50}{\makebox[0pt][l]{\hspace{-3mm}\strut \makecell[r]{Midnight-12k}}}&\rotatebox{50}{\makebox[0pt][l]{\hspace{-3mm}\strut \makecell[r]{Virchow2}}}&\rotatebox{50}{\makebox[0pt][l]{\hspace{-3mm}\strut \makecell[r]{Prov-GigaPath}}}&\rotatebox{50}{\makebox[0pt][l]{\hspace{-3mm}\strut \makecell[r]{UNI}}}&\rotatebox{50}{\makebox[0pt][l]{\hspace{-3mm}\strut \makecell[r]{Phikon-v2}}}\\%
\addlinespace[0.3cm]%
\toprule%
HEST{-}ccRCC&27.9&{-}&\textbf{29.5}&28.1&\underline{29.2}&20.9&27.2&24.9&25.1&27.4\\%
HEST{-}COAD&\textbf{34.9}&{-}&29.3&\underline{33.2}&30.9&31.8&25.8&32.0&25.8&25.6\\%
HEST{-}IDC&\textbf{62.7}&{-}&60.4&60.5&\underline{61.1}&59.9&59.7&58.1&58.5&56.8\\%
HEST{-}LUNG&\textbf{59.4}&{-}&58.1&57.4&57.5&\underline{58.3}&56.8&56.1&55.6&55.0\\%
HEST{-}LYMPH{-}IDC&\textbf{28.6}&{-}&26.2&27.4&26.6&\underline{27.5}&25.7&25.0&25.8&24.8\\%
HEST{-}PAAD&\textbf{54.2}&{-}&51.7&\underline{52.3}&51.1&50.2&47.8&49.9&48.8&47.7\\%
HEST{-}PRAD&\textbf{39.7}&{-}&38.6&37.5&36.2&37.1&35.3&\underline{39.1}&32.7&37.9\\%
HEST{-}RECTUM&\textbf{25.4}&{-}&22.6&22.7&\underline{24.0}&20.3&20.7&19.4&17.6&18.7\\%
HEST{-}SKCM&\textbf{70.0}&{-}&62.0&\underline{68.3}&66.1&64.8&64.0&59.7&62.3&58.4\\%
MSI CRC (patch)&\textbf{78.0}&{-}&\underline{72.6}&72.0&70.0&70.8&71.9&69.5&69.7&68.3\\%
MSI STAD (patch)&\textbf{76.1}&{-}&73.0&73.1&72.2&\underline{73.7}&72.5&68.1&68.6&69.2\\%
\addlinespace[0.3cm]%
HEST{-}Average&\textbf{44.8}&\underline{43.2}&42.0&43.0&42.5&41.2&40.3&40.5&39.1&39.1\\%
Molecular{-}Average&\textbf{50.6}&{-}&47.6&\underline{48.4}&47.7&46.8&46.1&45.6&44.6&44.5\\%
\addlinespace[0.3cm]%
\hline%
\addlinespace[0.3cm]%
BACH&91.7&\textbf{93.2}&\underline{92.3}&92.2&74.7&90.6&88.6&79.2&79.7&73.6\\%
BreakHis&\underline{88.3}&81.8&\textbf{89.9}&86.3&81.0&83.4&81.1&84.5&80.4&70.9\\%
CoNSeP&\textbf{66.1}&65.0&\underline{65.1}&63.6&64.5&62.4&64.5&63.0&63.0&62.9\\%
CRC{-}100k&\textbf{97.2}&96.8&\underline{96.9}&\underline{96.9}&96.4&96.7&96.6&95.9&96.1&95.3\\%
Gleason&\textbf{80.7}&78.5&78.4&77.6&77.3&\underline{79.0}&78.5&72.7&76.0&75.3\\%
MHIST&\textbf{88.5}&\underline{87.9}&86.8&82.6&85.1&81.2&86.5&83.3&83.8&79.6\\%
MoNuSAC&\underline{69.4}&\textbf{70.4}&67.4&64.5&68.1&65.8&66.8&66.4&65.0&64.3\\%
PANDA&\textbf{68.3}&\underline{66.6}&66.4&66.2&66.2&65.4&64.7&65.0&65.4&62.1\\%
PCAM&\textbf{96.1}&\underline{95.2}&\underline{95.2}&95.1&94.4&93.0&93.9&94.6&93.9&90.0\\%
CAMELYON16 (0.25 MPP)&\textbf{86.8}&{-}&\underline{86.3}&85.9&84.7&86.1&85.7&80.7&84.1&80.7\\%
TCGA Uniform (1.0 MPP)&\textbf{84.1}&{-}&78.7&79.4&76.0&\underline{83.7}&78.0&74.7&74.3&70.1\\%
TCGA Uniform (0.5 MPP)&\underline{82.7}&{-}&77.0&78.8&78.8&\textbf{83.2}&77.5&74.3&74.5&76.7\\%
\addlinespace[0.3cm]%
Morphology{-}Average{-}Pluto{-}4G&\textbf{82.9}&81.7&\underline{82.0}&80.6&78.6&79.7&80.1&78.3&78.1&74.9\\%
Morphology{-}Average&\textbf{83.3}&{-}&\underline{81.7}&80.8&78.9&80.9&80.2&77.9&78.0&75.1\\%
\addlinespace[0.3cm]%
\hline%
\addlinespace[0.3cm]%
Prediction{-}Average{-}Pluto{-}4G&\textbf{63.8}&\underline{62.5}&62.0&61.8&60.6&60.5&60.2&59.4&58.6&57.0\\%
Prediction{-}Average&\textbf{67.7}&{-}&\underline{65.4}&65.3&64.0&64.6&63.9&62.4&62.0&60.5\\%
\addlinespace[0.3cm]%
\hline%
\addlinespace[0.3cm]%
\addlinespace[0.3cm]%
\addlinespace[0.3cm]%
\hline%
\addlinespace[0.3cm]%
Robustness Index (Camelyon)&\textbf{94.0}&{-}&78.5&54.4&70.5&47.8&\underline{79.9}&38.2&14.5&1.9\\%
Robustness Index (TCGA)&\textbf{87.9}&{-}&82.6&80.3&81.2&\underline{85.8}&82.2&73.7&74.7&61.9\\%
Robustness Index (Tolkach)&\textbf{96.4}&{-}&93.8&92.3&91.8&94.1&\underline{95.4}&74.6&90.2&76.8\\%
Plismbench (Leaderboard metric)&\textbf{64.5}&{-}&41.1&36.0&\underline{52.2}&37.3&46.4&37.3&38.5&19.6\\%
\addlinespace[0.3cm]%
Robustness{-}Average&\textbf{85.7}&{-}&74.0&65.8&73.9&66.2&\underline{76.0}&56.0&54.5&40.0\\%
\addlinespace[0.3cm]%
\hline%
\addlinespace[0.1cm]%
\multicolumn{11}{r}{*Results for Pluto-4G taken from~\cite{pluto4g}.}
\end{tabular}
\end{adjustbox}
\vspace{3mm}
\caption{The three sections of the table show prediction results for tasks with morphology and molecular targets as well as for robustness benchmarks. The \textbf{best} result per row is in bold, the \underline{second-best} result is underlined. The presented model Atlas 2 exhibits best performance on the majority of the tasks across all three sections.}
\label{tab:main_results}
\end{table}

The following analysis is based on eighty public benchmark datasets from five public foundation model evaluation frameworks \textit{eva}~\cite{kaiko.ai2024eva}, HEST~\cite{jaume2024hest}, Plismbench~\cite{plismbench}, PathoROB~\cite{koemen2025pathorob}, and Patho-Bench~\cite{pathobench} as well as additional tumor microenvironment (TME) and MSI prediction benchmark datasets. The tasks range from TME tissue- and cell-typing over identifying morphological patterns, identifying cancer subtypes, to predicting gene mutations from histology. The evaluation protocols and task descriptions are detailed in Section~\ref{ssec:evaluation_protocols} and Appendix~\ref{ssec:task_descriptions}, respectively.

\begin{table}[p]
\centering
\begin{adjustbox}{width=1\textwidth}
\scriptsize
\begin{tabular}{l>{\raggedleft\arraybackslash}p{0.5cm}>{\raggedleft\arraybackslash}p{0.5cm}>{\raggedleft\arraybackslash}p{0.5cm}>{\raggedleft\arraybackslash}p{0.5cm}>{\raggedleft\arraybackslash}p{0.5cm}}%
&\rotatebox{50}{\makebox[0pt][l]{\hspace{-3mm}\strut \makecell[r]{Atlas 2}}}&\rotatebox{50}{\makebox[0pt][l]{\hspace{-3mm}\strut \makecell[r]{Virchow2}}}&\rotatebox{50}{\makebox[0pt][l]{\hspace{-3mm}\strut \makecell[r]{Atlas}}}&\rotatebox{50}{\makebox[0pt][l]{\hspace{-3mm}\strut \makecell[r]{UNI2-H}}}&\rotatebox{50}{\makebox[0pt][l]{\hspace{-3mm}\strut \makecell[r]{H-Optimus-0}}}\\%
\addlinespace[0.3cm]%
\toprule%
{[}CPTAC BRCA: Breast{]}{[}Macro OvR AUC{]} Mutation prediction: TP53&\textbf{83.6}&\underline{78.7}&78.2&77.7&76.0\\%
{[}CPTAC BRCA: Breast{]}{[}Macro OvR AUC{]} Mutation prediction: PIK3CA&59.6&58.2&59.4&\textbf{61.1}&\underline{59.9}\\%
{[}CPTAC CCRCC: Kidney{]}{[}Macro OvR AUC{]} Mutation prediction: BAP1&\underline{70.4}&67.4&\textbf{70.5}&66.9&65.3\\%
{[}CPTAC CCRCC: Kidney{]}{[}Macro OvR AUC{]} Mutation prediction: PBRM1&\textbf{52.4}&\underline{49.4}&48.8&41.6&44.0\\%
{[}CPTAC CCRCC: Kidney{]}{[}Macro OvR AUC{]} Mutation prediction: VHL&\textbf{52.9}&\underline{52.8}&48.7&48.2&49.7\\%
{[}CPTAC COAD: Colon{]}{[}Macro OvR AUC{]} Mutation prediction: KRAS&\underline{65.3}&64.3&61.7&\textbf{66.4}&60.4\\%
{[}CPTAC COAD: Colon{]}{[}Macro OvR AUC{]} Mutation prediction: ACVR2A&\textbf{84.5}&78.2&76.9&75.7&\underline{79.5}\\%
{[}CPTAC COAD: Colon{]}{[}Macro OvR AUC{]} Mutation prediction: APC&\textbf{77.8}&72.9&\underline{74.7}&73.8&71.5\\%
{[}CPTAC COAD: Colon{]}{[}Macro OvR AUC{]} Mutation prediction: ARID1A&\textbf{75.8}&71.1&74.0&69.7&\underline{75.6}\\%
{[}CPTAC COAD: Colon{]}{[}Macro OvR AUC{]} Mutation prediction: MSI\_H&\textbf{93.9}&85.9&84.6&\underline{89.0}&87.1\\%
{[}CPTAC COAD: Colon{]}{[}Macro OvR AUC{]} Mutation prediction: PIK3CA&62.2&\underline{63.7}&57.1&59.1&\textbf{63.9}\\%
{[}CPTAC COAD: Colon{]}{[}Macro OvR AUC{]} Mutation prediction: SETD1B&\textbf{82.3}&75.3&76.0&79.9&\underline{81.8}\\%
{[}CPTAC COAD: Colon{]}{[}Macro OvR AUC{]} Mutation prediction: TP53&\textbf{72.5}&66.1&62.3&\underline{67.3}&67.1\\%
{[}CPTAC GBM: Brain{]}{[}Macro OvR AUC{]} Mutation prediction: EGFR&\textbf{69.2}&\underline{62.1}&59.1&57.5&58.0\\%
{[}CPTAC GBM: Brain{]}{[}Macro OvR AUC{]} Mutation prediction: TP53&\textbf{85.5}&74.8&\underline{83.7}&81.9&81.8\\%
{[}CPTAC HNSC: Head\&Neck{]}{[}Macro OvR AUC{]} Mutation prediction: CASP8&\textbf{65.8}&56.6&57.6&\underline{60.1}&54.9\\%
{[}CPTAC LSCC: Lung{]}{[}Macro OvR AUC{]} Mutation prediction: KEAP1&57.4&63.1&\underline{63.4}&\textbf{65.0}&62.6\\%
{[}CPTAC LSCC: Lung{]}{[}Macro OvR AUC{]} Mutation prediction: ARID1A&\textbf{46.9}&41.9&42.6&\underline{43.4}&41.8\\%
{[}CPTAC LUAD: Lung{]}{[}Macro OvR AUC{]} Mutation prediction: EGFR&81.0&\textbf{81.7}&\underline{81.1}&77.1&77.8\\%
{[}CPTAC LUAD: Lung{]}{[}Macro OvR AUC{]} Mutation prediction: KRAS&70.4&\textbf{73.9}&66.9&65.4&\underline{72.1}\\%
{[}CPTAC LUAD: Lung{]}{[}Macro OvR AUC{]} Mutation prediction: STK11&\underline{88.5}&82.4&85.6&\textbf{88.8}&88.4\\%
{[}CPTAC LUAD: Lung{]}{[}Macro OvR AUC{]} Mutation prediction: TP53&\underline{76.7}&\textbf{77.3}&69.6&76.0&71.4\\%
{[}CPTAC PDA: Pancreas{]}{[}Macro OvR AUC{]} Mutation prediction: SMAD4&\textbf{47.6}&44.6&\underline{45.1}&37.0&41.4\\%
{[}BC Therapy: Breast{]}{[}Macro OvR AUC{]} Molecular Subtyping: HER2 status&\textbf{69.9}&\underline{64.3}&60.9&63.6&61.2\\%
\addlinespace[0.3cm]%
Molecular{-}Average&\textbf{70.5}&\underline{66.9}&66.2&66.3&66.4\\%
\addlinespace[0.3cm]%
\hline%
\addlinespace[0.3cm]%
{[}CPTAC BRCA: Breast{]}{[}Balanced Accuracy{]} TME Characterization: Immune class&\textbf{59.3}&\underline{55.8}&51.7&53.8&52.2\\%
{[}CPTAC CCRCC: Kidney{]}{[}Balanced Accuracy{]} TME Characterization: Immune class&\textbf{38.3}&33.1&\underline{35.7}&32.1&34.7\\%
{[}CPTAC COAD: Colon{]}{[}Balanced Accuracy{]} TME Characterization: Immune class&\textbf{45.8}&\underline{42.1}&39.1&38.4&36.0\\%
{[}CPTAC GBM: Brain{]}{[}Balanced Accuracy{]} TME Characterization: Immune class&\textbf{57.7}&46.9&\underline{52.0}&51.2&50.9\\%
{[}CPTAC HNSC: Head\&Neck{]}{[}Balanced Accuracy{]} TME Characterization: Immune class&\textbf{54.6}&\underline{51.5}&48.6&46.9&48.6\\%
{[}CPTAC PDA: Pancreas{]}{[}Balanced Accuracy{]} TME Characterization: Immune class&\underline{41.7}&\textbf{43.0}&40.8&39.2&39.2\\%
{[}CPTAC LSCC: Lung{]}{[}Balanced Accuracy{]} TME Characterization: Immune class&\textbf{57.8}&\underline{57.2}&54.2&50.7&53.7\\%
{[}CPTAC LUAD: Lung{]}{[}Balanced Accuracy{]} TME Characterization: Immune class&\textbf{56.6}&50.2&\underline{50.3}&47.3&46.8\\%
{[}Hancock: Head\&Neck{]}{[}Macro OvR AUC{]} TME Characterization: Lymphovascular invasion detection&\textbf{69.9}&64.8&\underline{69.3}&65.4&68.0\\%
{[}Hancock: Head\&Neck{]}{[}Macro OvR AUC{]} TME Characterization: Perineural invasion detection&\textbf{74.0}&72.3&71.8&68.5&\underline{72.5}\\%
{[}Hancock: Head\&Neck{]}{[}Macro OvR AUC{]} TME Characterization: Primary vs. Metastasis&\textbf{65.8}&65.0&61.9&\underline{65.2}&62.6\\%
{[}Hancock: Head\&Neck{]}{[}Balanced Accuracy{]} TME Characterization: Primary tumor site&\textbf{77.5}&\underline{74.6}&74.1&73.9&74.4\\%
{[}Hancock: Head\&Neck{]}{[}Macro OvR AUC{]} TME Characterization: Vascular invasion detection&67.6&63.0&66.9&\underline{68.2}&\textbf{70.4}\\%
{[}BC Therapy: Breast{]}{[}Weighted Kappa{]} Tumor Grading&\textbf{39.5}&32.4&29.8&\underline{34.9}&27.8\\%
{[}CPTAC HNSC: Head\&Neck{]}{[}Weighted Kappa{]} Tumor Grading&\underline{36.1}&36.0&29.8&\textbf{36.6}&33.6\\%
{[}CPTAC LSCC: Lung{]}{[}Weighted Kappa{]} Tumor Grading&22.7&22.1&23.8&\textbf{26.7}&\underline{25.1}\\%
{[}Hancock: Head\&Neck{]}{[}Weighted Kappa{]} Tumor Grading: Conventional Keratinizing&\textbf{33.2}&28.0&24.7&29.1&\underline{30.9}\\%
{[}Hancock: Head\&Neck{]}{[}Weighted Kappa{]} Tumor Grading: Conventional NonKeratinizing&\textbf{23.7}&20.4&20.6&\underline{21.0}&18.1\\%
{[}IMP Cervical: Cervix{]}{[}Macro OvR AUC{]} Morphological subtyping&94.7&95.1&95.0&\underline{95.4}&\textbf{95.5}\\%
\addlinespace[0.3cm]%
Morphology{-}Average&\textbf{53.5}&\underline{50.2}&49.5&49.7&49.5\\%
\addlinespace[0.3cm]%
\hline%
\addlinespace[0.3cm]%
{[}Boehmk: Ovary{]}{[}C{-}Index{]} Survival prediction&\textbf{51.7}&50.8&49.5&\underline{50.9}&49.2\\%
{[}CPTAC CCRCC: Kidney{]}{[}C{-}Index{]} Survival prediction&55.6&58.2&\textbf{65.4}&\underline{63.8}&55.2\\%
{[}CPTAC HNSC: Head\&Neck{]}{[}C{-}Index{]} Survival prediction&51.1&50.8&48.3&\underline{56.7}&\textbf{60.2}\\%
{[}CPTAC PDA: Pancreas{]}{[}C{-}Index{]} Survival prediction&\textbf{52.0}&50.7&50.7&49.9&\underline{51.3}\\%
{[}CPTAC LUAD: Lung{]}{[}C{-}Index{]} Survival prediction&52.2&\textbf{57.1}&52.5&48.1&\underline{54.2}\\%
{[}MBC: Breast{]}{[}C{-}Index{]} Survival prediction&49.0&51.8&\textbf{54.2}&51.0&\underline{53.0}\\%
{[}Hancock: Head\&Neck{]}{[}C{-}Index{]} Survival prediction&\textbf{57.0}&56.4&56.5&53.9&\underline{56.7}\\%
\addlinespace[0.3cm]%
Survival{-}Average&52.7&53.7&\underline{53.9}&53.5&\textbf{54.3}\\%
\addlinespace[0.3cm]%
\hline%
\addlinespace[0.3cm]%
{[}BC Therapy: Breast{]}{[}Macro OvR AUC{]} Treatment response: ER status&\textbf{76.4}&66.1&69.2&\underline{70.0}&60.5\\%
{[}BC Therapy: Breast{]}{[}Balanced Accuracy{]} Treatment Response: Residual cancer burden&\textbf{33.7}&29.0&29.5&\underline{31.2}&27.4\\%
{[}MBC: Breast{]}{[}Weighted Kappa{]} Treatment Response: Recist&10.3&8.5&\textbf{16.1}&10.2&\underline{12.1}\\%
\addlinespace[0.3cm]%
Treatment{-}Response{-}Average&\textbf{40.1}&34.5&\underline{38.3}&37.1&33.3\\%
\addlinespace[0.3cm]%
\hline%
\addlinespace[0.3cm]%
Prediction{-}Average&\textbf{60.3}&\underline{57.4}&57.0&57.0&56.9\\%
\addlinespace[0.3cm]%
\hline%
\end{tabular}

\end{adjustbox}
\vspace{3mm}
\caption{The four sections of the table show prediction results for tasks with morphology, molecular, survival, and treatment targets from the Patho-Bench~\cite{pathobench} framework. The \textbf{best} result per row is in bold, the \underline{second-best} result is underlined. The presented model Atlas 2 exhibits best performance on the majority of the tasks across three sections. For survival prediction H-Optimus-0~\cite{hoptimus0} shows best performance, but it is to note that all models perform close to chance level (C-Index 50).}
\label{tab:main_results_pathobench}
\end{table}

\begin{table}[!ht]
\centering
\begin{adjustbox}{width=1\textwidth}
\scriptsize
\begin{tabular}{l>{\raggedleft\arraybackslash}p{0.5cm}@{\hspace{5mm}\vrule}>{\raggedleft\arraybackslash}p{0.5cm}>{\raggedleft\arraybackslash}p{0.5cm}>{\raggedleft\arraybackslash}p{0.5cm}>{\raggedleft\arraybackslash}p{0.5cm}>{\raggedleft\arraybackslash}p{0.5cm}@{\hspace{5mm}\vrule}>{\raggedleft\arraybackslash}p{0.5cm}>{\raggedleft\arraybackslash}p{0.5cm}>{\raggedleft\arraybackslash}p{0.5cm}>{\raggedleft\arraybackslash}p{0.5cm}>{\raggedleft\arraybackslash}p{0.5cm}}%
&&\multicolumn{5}{c|}{\textbf{ViT-B}}&\multicolumn{5}{c}{\textbf{ViT-S}}\\[0.8cm]%
&\rotatebox{50}{\makebox[0pt][l]{\hspace{-3mm}\strut \makecell[r]{Atlas 2}}}&\rotatebox{50}{\makebox[0pt][l]{\hspace{-3mm}\strut \makecell[r]{Atlas 2-B}}}&\rotatebox{50}{\makebox[0pt][l]{\hspace{-3mm}\strut \makecell[r]{ViT-B/8 \cite{kaiko2024towardslarge}}}}&\rotatebox{50}{\makebox[0pt][l]{\hspace{-3mm}\strut \makecell[r]{Hibou-B}}}&\rotatebox{50}{\makebox[0pt][l]{\hspace{-3mm}\strut \makecell[r]{Phikon}}}&\rotatebox{50}{\makebox[0pt][l]{\hspace{-3mm}\strut \makecell[r]{H0-mini*}}}&\rotatebox{50}{\makebox[0pt][l]{\hspace{-3mm}\strut \makecell[r]{Atlas 2-S}}}&\rotatebox{50}{\makebox[0pt][l]{\hspace{-3mm}\strut \makecell[r]{Pluto-4S-8*}}}&\rotatebox{50}{\makebox[0pt][l]{\hspace{-3mm}\strut \makecell[r]{Pluto-4S-16*}}}&\rotatebox{50}{\makebox[0pt][l]{\hspace{-3mm}\strut \makecell[r]{ViT-S/8 \cite{kang2022benchmarking}}}}&\rotatebox{50}{\makebox[0pt][l]{\hspace{-3mm}\strut \makecell[r]{ViT-S/8 \cite{kaiko2024towardslarge}}}}\\[8pt]%
\toprule%
HEST{-}ccRCC&27.9&\textbf{26.8}&24.8&22.8&24.2&\underline{26.4}&\underline{26.4}&{-}&{-}&\textbf{26.7}&25.7\\%
HEST{-}COAD&34.9&\textbf{32.4}&\underline{30.2}&28.7&27.8&27.0&\textbf{32.0}&{-}&{-}&\underline{30.2}&23.0\\%
HEST{-}IDC&62.7&\textbf{61.2}&57.0&56.0&54.8&\underline{59.1}&\textbf{60.8}&{-}&{-}&\underline{55.2}&54.5\\%
HEST{-}LUNG&59.4&\textbf{56.8}&53.9&54.9&\underline{56.7}&56.3&\textbf{54.9}&{-}&{-}&\underline{54.4}&53.2\\%
HEST{-}LYMPH{-}IDC&28.6&\textbf{27.1}&23.5&25.0&23.8&\underline{26.4}&\textbf{26.6}&{-}&{-}&\underline{24.9}&23.9\\%
HEST{-}PAAD&54.2&\textbf{52.0}&47.8&45.9&46.5&\underline{50.7}&\textbf{49.7}&{-}&{-}&\underline{44.2}&43.3\\%
HEST{-}PRAD&39.7&\textbf{38.8}&\underline{38.0}&30.6&34.5&36.3&\textbf{35.7}&{-}&{-}&27.1&\underline{34.4}\\%
HEST{-}RECTUM&25.4&\textbf{24.5}&17.0&17.6&16.6&\underline{20.5}&\textbf{20.8}&{-}&{-}&\underline{15.5}&14.4\\%
HEST{-}SKCM&70.0&\textbf{63.4}&59.4&55.9&54.5&\underline{61.2}&\textbf{60.8}&{-}&{-}&\underline{57.5}&56.5\\%
MSI CRC (patch)&78.0&\textbf{74.9}&\underline{68.8}&68.3&67.2&{-}&\textbf{74.9}&{-}&{-}&\underline{70.2}&69.4\\%
MSI STAD (patch)&76.1&\textbf{75.0}&68.1&\underline{71.0}&67.1&{-}&\textbf{78.1}&{-}&{-}&\underline{72.6}&72.3\\%
\addlinespace[0.3cm]%
HEST{-}Average&44.8&\textbf{42.6}&39.1&37.5&37.7&\underline{40.4}&\textbf{40.9}&36.9&36.4&\underline{37.3}&36.5\\%
Molecular{-}Average&50.6&\textbf{48.4}&\underline{44.4}&43.3&43.1&{-}&\textbf{47.3}&{-}&{-}&\underline{43.5}&42.8\\%
\addlinespace[0.3cm]%
\hline%
\addlinespace[0.3cm]%
BACH&91.7&\textbf{90.9}&\underline{88.0}&81.9&73.5&{-}&\textbf{90.4}&\underline{85.1}&79.8&77.3&80.2\\%
BreakHis&88.3&\textbf{90.9}&\underline{84.4}&79.8&71.0&{-}&\textbf{84.0}&\underline{80.8}&76.8&72.7&73.0\\%
CoNSeP&66.1&\textbf{65.8}&\underline{64.5}&63.3&62.8&62.9&\textbf{65.9}&\underline{64.9}&62.1&63.4&63.9\\%
CRC{-}100k&97.2&\textbf{97.0}&\underline{96.0}&95.8&94.7&{-}&\textbf{97.1}&95.4&\underline{95.5}&94.9&95.1\\%
Gleason&80.7&\textbf{78.7}&\underline{75.4}&73.6&74.2&{-}&\textbf{80.5}&76.3&\underline{76.6}&74.3&73.3\\%
MHIST&88.5&\textbf{88.0}&\underline{83.0}&79.9&82.3&{-}&\textbf{84.7}&\underline{84.2}&83.4&76.5&82.7\\%
MoNuSAC&69.4&\textbf{69.3}&\underline{68.5}&64.9&64.3&64.3&\underline{67.2}&\textbf{67.8}&64.0&66.7&65.3\\%
PANDA&68.3&\textbf{68.4}&\underline{64.7}&63.2&64.5&{-}&\textbf{67.2}&\underline{63.1}&63.0&61.9&60.3\\%
PCAM&96.1&\textbf{95.6}&92.0&\underline{94.3}&92.8&{-}&\textbf{94.5}&91.4&91.6&\underline{92.4}&88.6\\%
CAMELYON16 (0.25 MPP)&86.8&\textbf{87.5}&81.9&80.3&\underline{82.8}&{-}&\textbf{86.0}&{-}&{-}&79.1&\underline{81.1}\\%
TCGA Uniform (1.0 MPP)&84.1&\textbf{81.4}&\underline{80.5}&72.4&70.8&{-}&\textbf{77.0}&{-}&{-}&71.9&\underline{74.0}\\%
TCGA Uniform (0.5 MPP)&82.7&\textbf{79.5}&\underline{79.2}&71.1&75.6&{-}&\textbf{75.6}&{-}&{-}&71.5&\underline{72.2}\\%
\addlinespace[0.3cm]%
Morphology{-}Average{-}Pluto{-}4G&82.9&\textbf{82.7}&\underline{79.6}&77.4&75.6&{-}&\textbf{81.3}&\underline{78.8}&77.0&75.6&75.8\\%
Morphology{-}Average&83.3&\textbf{82.8}&\underline{79.8}&76.7&75.8&{-}&\textbf{80.8}&{-}&{-}&75.2&\underline{75.8}\\%
\addlinespace[0.3cm]%
\hline%
\addlinespace[0.3cm]%
Prediction{-}Average{-}Pluto{-}4G&63.8&\textbf{62.6}&\underline{59.3}&57.4&56.6&{-}&\textbf{61.1}&\underline{57.8}&56.7&56.4&56.2\\%
Prediction{-}Average&67.7&\textbf{66.3}&\underline{62.9}&60.7&60.1&{-}&\textbf{64.8}&{-}&{-}&\underline{60.0}&\underline{60.0}\\%
\addlinespace[0.3cm]%
\hline%
\addlinespace[0.3cm]%
\addlinespace[0.3cm]%
\addlinespace[0.3cm]%
\hline%
\addlinespace[0.3cm]%
Robustness Index (Camelyon)&94.0&\textbf{95.3}&14.7&5.2&1.1&\underline{71.8}&\textbf{95.6}&{-}&{-}&4.3&\underline{26.5}\\%
Robustness Index (TCGA)&87.9&\textbf{88.2}&76.3&61.3&62.3&\underline{79.4}&\textbf{87.5}&{-}&{-}&66.1&\underline{74.0}\\%
Robustness Index (Tolkach)&96.4&\textbf{95.9}&89.6&79.5&79.5&\underline{93.2}&\textbf{96.3}&{-}&{-}&83.2&\underline{91.8}\\%
Plismbench (Leaderboard metric)&64.5&\textbf{65.3}&\underline{35.8}&24.6&24.4&{-}&\textbf{64.1}&{-}&{-}&32.1&\underline{37.4}\\%
\addlinespace[0.3cm]%
Robustness{-}Average&85.7&\textbf{86.2}&\underline{54.1}&42.6&41.8&{-}&\textbf{85.9}&{-}&{-}&46.4&\underline{57.4}\\%
\addlinespace[0.3cm]%
\hline%
\addlinespace[0.1cm]%
\multicolumn{12}{r}{*Results for H0-mini taken from~\cite{plismbench} (HEST, \textit{eva}) and \cite{pathorobgithub} (PathoROB). Results for Pluto-4S-8 and Pluto-4S-16 taken from~\cite{pluto4g}.}
\end{tabular}%

\end{adjustbox}
\vspace{3mm}
\caption{Comparison of resource efficient pathology foundation models for the same benchmarks as in Table~\ref{tab:main_results}. To better compare models of similar sizes, we highlight the \textbf{best} and \underline{second-best} results per task for each architecture (ViT-B and ViT-S) separately. Atlas 2-B and 2-S exhibit best performance in their respective architecture category on the majority of tasks.}
\label{tab:distillation_results}
\end{table}

\subsection{Comparison of large-scale pathology foundation models}

Table~\ref{tab:main_results} shows that Atlas 2 achieves the best performance in 22/27 tasks and second-best performance in 3 additional tasks with an average performance score of 44.8\% on HEST, a 1.6 p.p.\ improvement compared to the second-best performing model Pluto-4G~\cite{pluto4g}, 82.9\% on \textit{eva}\footnote{For comparability this considers the benchmark subset of \textit{eva} evaluated in~\cite{pluto4g}.}, a 1.2 p.p.\ improvement over the closest contender Pluto-4G~\cite{pluto4g}, and an average robustness score of 85.7\%, a 9.7 p.p.\ improvement to the closest contender Virchow2~\cite{zimmermann_virchow_2024}. Only a limited comparison to the recent model H-Optimus-1~\cite{hoptimus1} is possible. The results in Table~\ref{tab:main_results_cls} demonstrate that Atlas 2 performs better than H-Optimus-1~\cite{hoptimus1} on the shared benchmarks HEST, CRC-100k, and MHIST with an improvement of 1, 1.5, and 4.5 p.p., respectively.

In Table~\ref{tab:main_results_pathobench} we report additional analyses for molecular and morphology related tasks using the Patho-Bench framework~\cite{pathobench}. Results demonstrate that Atlas 2 is the best-performing model in 15/24 molecular related tasks and achieves at least second-best performance in 19/24 molecular related tasks with an average improvement of 3.6 p.p.\ compared to the second-best model, Virchow2. For morphology related benchmarks, Atlas 2 is the top performing model in 14/19 tasks and achieves at least second-best performance in 16/19 tasks with an average difference of 3.3 p.p.\ compared to the second-best model. Similarly, Atlas 2 shows the highest performance in treatment response prediction tasks with a 1.8 p.p.\ improvement compared to the second-best performing model. For survival prediction tasks, H-Optimus-0~\cite{hoptimus0} is the best-performing model, but it is noteworthy that all models perform close to the chance level (C-Index of 50) for this task.

\subsection{Comparison of resource efficient pathology foundation models}

Table~\ref{tab:distillation_results} shows results for resource efficient models.
Atlas 2-B and Atlas 2-S are distilled from Atlas~2 and are 24 and 91 times smaller in parameter size and 3.4 and 9 times more resource efficient for a standard-size image inference.
The results show that Atlas 2-B and Atlas 2-S are the best-performing models in 27/27 and 25/27 tasks, respectively, within their compute category. Atlas 2-B performs on par with Pluto-4G~\cite{pluto4g} while running standard-size image inference more than twice as fast (174 vs.\ 79 img/s) and outperforms the larger models Atlas~\cite{alber2025atlas} and UNI2-H~\cite{chen_towards_2024} (66.3 vs.\ 65.4 and 65.3).
Atlas 2-S outperforms the larger models H-Optimus-0~\cite{hoptimus0} and Virchow2~\cite{zimmermann_virchow_2024} (64.8 vs.\ 64.0 and 63.9) despite being 6.6 and 4.3 times more resource efficient for a standard-size image inference, respectively.
The models also show remarkable robustness: Atlas 2-B and Atlas 2-S are even more robust than Atlas 2, reaching a robustness average of 86.2 and 85.9, respectively.
Due to a lack of reference values and model access, H0-mini~\cite{filiot2025distillingfoundationmodelsrobust} could not be compared on every robustness benchmark.
A comparison to H0-mini~\cite{filiot2025distillingfoundationmodelsrobust} on published results shows that, for the CLS+MEAN representation, the average robustness on PathoROB~\cite{koemen2025pathorob} is 11.6 p.p.\ better for Atlas 2-B (93.1 vs.\ 81.5).
On the CLS token evaluation for Plismbench~\cite{plismbench}, Atlas 2-B outperforms H0-mini by 4.4 p.p.\ (58.5 vs.\ 54.1).

\section{Discussion}
For clinical deployment, foundation models must demonstrate both high prediction performance for the task at hand and robustness to confounding data variations commonly seen in routine diagnostics. 
Additionally, the large volume of histological slides produced in pathology labs necessitates efficient processing to provide timely and cost-effective analyses. 
Our novel pathology foundation models Atlas 2, as well as its efficient, distilled versions 2-B and 2-S, were trained on the largest dataset to date used for pathology foundation model development comprising 5.5 million histological slides. 
Our models show consistent state-of-the-art performance on all three important requirements --- prediction performance, robustness, and resource efficiency. 
The evaluation was performed on eighty benchmarks from five different public benchmark suites covering a wide variety of tasks, datasets, and tissue types. 
These are the properties that make foundation models useful in practice: applications built on our Atlas family, such as Atlas H\&E-TME~\cite{standvoss2026atlas} for expert-level tumor microenvironment profiling from routine H\&E and the derived OpenTME~\cite{galama2026opentme} dataset of cell-level profiles across thousands of TCGA slides, translate these capabilities into scalable, quantitative tissue analysis. 
While further improvements and benchmarking are warranted, our models outperform all other evaluated pathology foundation models on the vast majority of benchmarks and demonstrate that tradeoffs for clinical deployments of pathology foundation models can effectively be minimized. 

\section*{Acknowledgments}
We would like to thank the teams at Aignostics, Charité - Universitätsmedizin Berlin - Pathology Department, LMU Munich - Pathology Department, Mayo Clinic Digital Pathology, Mayo Clinic Generative Artificial Intelligence Program, Mayo HPC team, and Mayo Clinic Department of Laboratory Medicine and Pathology for supporting this work. 
Special thanks go to Edwin de Jong, Pavel Trunov, Paul Mattes, Evelyn Ramberger, Hans Pinckaers, Sergey Melnik, Steele Clifton-Berry, Viktor Matyas, Crystal D. Liudahl, Zach D. Jensen, and Wyman D. Matthews for their organizational, technical, and other support.

The benchmark results shown here are in part based upon data generated by the TCGA Research Network: \url{https://www.cancer.gov/tcga}.

This work was in part supported by the German Ministry for Education and Research (BMBF) under Grants 01IS14013A-E, 01GQ1115, 01GQ0850, 01IS18025A,
031L0207D, and 01IS18037A.  K.R.M. was partly supported by the Institute of Information \& Communications Technology Planning \& Evaluation (IITP) grants funded by the Korea government (MSIT) (No. 2019-0-00079, Artificial Intelligence Graduate School Program, Korea University and No. 2022-0-00984, Development of Artificial Intelligence Technology for Personalized Plug-and-Play Explanation and Verification of Explanation).


\bibliographystyle{plain} 
\bibliography{references} 

\newpage
\appendix

\section{Appendix}

\subsection{Task Descriptions}
\label{ssec:task_descriptions}

\paragraph{HEST-Benchmark} The HEST-Benchmark was introduced by~\cite{jaume2024hest} for benchmarking foundation models for histology on the task of gene expression prediction from H\&E-stained images. The benchmark includes 72 spatial transcriptomics profiles (using Xenium or Visium technology) with corresponding H\&E-stained images from 47 patients grouped into 10 tasks based on organ. Each task involves predicting the expression levels of the 50 most variable genes (highest normalized variance) from 112$\times$112 {\textmu}m H\&E-stained image patches (equivalent to 224$\times$224 pixels at 20$\times$ magnification) centered on each spatial transcriptomics spot, formulated as a multivariate regression problem. We use the default Ridge Regression with PCA (256 factors) evaluation protocol to solve the multivariate regression problem for extracted embeddings~\cite{jaume2024hest}. Specifically, the objective of the 10 tasks is to predict gene expression levels for invasive ductal carcinoma (breast cancer, IDC), prostate adenocarcinoma (prostate cancer, PRAD), pancreatic adenocarcinoma (pancreatic cancer, PAAD), skin cutaneous melanoma (skin cancer, SKCM), colonic adenocarcinoma (colon cancer, COAD), rectal adenocarcinoma (rectum cancer, READ), clear cell renal cell carcinoma (kidney cancer, ccRCC), hepatocellular carcinoma (liver cancer, HCC), lung adenocarcinoma (lung cancer, LUAD), and axillary lymph nodes in IDC (metastatic, LYMPH-IDC). The benchmark applies patient-stratified splits to avoid any train/test patient-level data leakage, resulting in a k-fold cross-validation where k is the number of patients~\cite{jaume2024hest}. Performance is measured as the Pearson correlation coefficient between the predicted and measured gene expression. Results are reported as the mean across folds.

\paragraph{BACH} The BACH dataset comprises 400 H\&E microscopy images (2048$\times$1536 pixels at 20$\times$ magnification) of breast cancer biopsies. It was introduced as part of the ICIAR 2018 Grand Challenge on BreAst Cancer Histology images (BACH)~\cite{aresta2019bach}. The task of the challenge is to classify each image into one of the following four classes: ``normal'', ``benign'', ``in situ carcinoma'', and ``invasive carcinoma''. The dataset is split into 268 training images (67 images per class) and 132 test images (33 images per class). The patches are resized and cropped to 224$\times$224 pixels.

\paragraph{CRC-100k} The CRC-100k dataset contains 107,180 H\&E images (224$\times$224 pixels at 20$\times$ magnification) extracted from colorectal cancer (CRC) tissue samples~\cite{kather2019predicting}. The tissue samples originate from CRC primary tumors and CRC liver metastases. The task of this benchmark is to classify each image into one of the following 9 tissue classes: ``adipose'', ``background'', ``debris'', ``lymphocytes'', ``mucus'', ``smooth muscle'', ``normal colon mucosa'', ``cancer-associated stroma'', and ``colorectal adenocarcinoma epithelium''. The dataset is split into 100,000 training images (NCT-CRC-HE-100K-NONORM) and 7,180 test images (CRC-VAL-HE-7K). We use the original (no-norm) images without Macenko color normalization~\cite{crc-100k}.

\paragraph{MHIST} The task of MHIST is to classify images of colorectal polyps into hyperplastic polyps (HPs) vs.\ sessile serrated adenomas (SSAs)~\cite{mhist}. This distinction is clinically important as HPs are typically benign whereas SSAs are precancerous lesions that can turn into cancer if left untreated. The task is challenging for pathologists, often showing considerable inter-pathologist variability. The MHIST dataset consists of 3,152 H\&E images (224$\times$224 pixels at 8$\times$ magnification) of colorectal polyps. Labels are derived from the majority vote of seven pathologists. The dataset is split into 2,162 training images (1,545 HP and 617 SSA) and 990 test images (630 HP and 360 SSA). 

\paragraph{PCAM} PCAM (PatchCamelyon) defines the clinically-relevant task of metastasis detection as a binary image classification task~\cite{pcam}. The dataset consists of 327,680 H\&E images (96$\times$96 pixels at 10$\times$ magnification) extracted from scans of sentinel lymph node sections. Each image is annotated with a binary label indicating the presence of metastatic breast cancer tissue. An image is labeled as metastatic if the center (32$\times$32 pixels) region contains at least one pixel of tumor tissue. The dataset is split by ratios of 80:10:10 into training, validation, and test sets with no overlap of WSIs/cases between the splits. Moreover, every split has a 50:50 balance of positive and negative examples. For evaluation, we resize each image to 224$\times$224 pixels. PCAM has been derived from the CAMELYON16 Challenge~\cite{camelyon}.

\paragraph{CAMELYON16} The task of the CAMELYON16 challenge~\cite{camelyon} is to classify whole slide images (WSIs) of lymph node tissue sections into metastatic breast cancer tissue being present or not. The dataset comprises 399 H\&E-stained WSIs of sentinel lymph node sections, which were acquired and scanned (40$\times$ magnification) at two centers from the Netherlands~\cite{camelyon}. The dataset is split into 270 training slides (110 with and 160 without metastasis) and 129 test slides (49 with and 80 without metastasis). Here, we report results for the CAMELYON16 (small) setup in \textit{eva}~\cite{kaiko.ai2024eva}, which randomly samples max.\ 1,000 patches per slide.

\paragraph{PANDA} The PANDA challenge~\cite{panda} is concerned with the challenging task of tumor grading of whole slide images (WSIs) of prostate cancer biopsies, which suffers from significant inter-observer variability between pathologists. Prostate cancer grading follows the Gleason grading system (3, 4, or 5) based on architectural growth patterns of the tumor, which are then converted into an ISUP grade on a scale of 1--5 for usage as a prognostic marker. The dataset features 9,555 H\&E-stained WSIs (subset with noisy labels removed) of prostate tissue biopsies from two medical centers scanned at 20$\times$ magnification. Specifically, the task is to classify each WSI into an ISUP grade of 0--5 (six classes), where 0 means that a biopsy does not contain any cancer. The dataset is split into 6,686 training slides, 1,430 validation slides, and 1,439 test slides in a class-stratified manner. Here, we report results for the PANDA (small) setup in \textit{eva}~\cite{kaiko.ai2024eva}, which considers a fewer number of total slides (955 train, 477 validation, 477 test) and fewer randomly sampled patches per slide (200).

\paragraph{BreakHis} The Breast Cancer Histopathological Image Classification (BreakHis) dataset was introduced for the task of classifying breast tumor tissue into benign and malignant histological subtypes~\cite{breakhis}. The task involves multi-class classification of microscopic images into eight categories: ``adenosis'' (A, benign), ``fibroadenoma'' (F, benign), ``phyllodes tumor'' (PT, benign), ``tubular adenoma'' (TA), ``carcinoma'' (DC, malignant), ``lobular carcinoma'' (LC, malignant), ``mucinous carcinoma'' (MC, malignant) and ``papillary carcinoma'' (PC, malignant). 
The original dataset consists of 9,109 microscopic images collected from 82 patients at multiple magnification factors, 40$\times$, 100$\times$, 200$\times$, and 400$\times$. For this benchmark, we follow the \textit{eva}~\cite{kaiko.ai2024eva} evaluation protocol and use only the 40$\times$ magnification subset, which consists of 1,995 H\&E-stained images (700$\times$460 pixels at 40$\times$ magnification, 0.49 {\textmu}m/pixel) selected from classes with at least 7 patients to ensure statistical robustness: TA, MC, F \& DC. The dataset is split using patient-stratified partitioning, resulting in 1,132 training images and 339 validation images with approximate class balance maintained across splits.

\paragraph{CoNSeP} The CoNSeP (Colorectal Nuclear Segmentation and Phenotypes) dataset was introduced for the task of nuclear segmentation and classification in colorectal tissue~\cite{consep}. The task involves segmenting and classifying cell nuclei into four categories: ``miscellaneous'', ``inflammatory'', ``epithelial'' (combining normal and malignant/dysplastic epithelial cells), and ``spindle-shaped'' (combining fibroblast, muscle, and endothelial cells). 
The dataset consists of 41 H\&E-stained images (1,000$\times$1,000 pixels at 40$\times$ magnification) extracted from 16 whole-slide images of unique colorectal cancer patients, containing 24,319 annotated nuclei in total. Segmentation masks are provided as instance-level annotations indicating both the spatial extent and class label of each nucleus. The dataset is split into 27 training images and 14 test images. Following the \textit{eva}~\cite{kaiko.ai2024eva} protocol, we use the test split as the validation split.

\paragraph{Gleason Arvaniti} The Gleason Arvaniti dataset was introduced for automated Gleason grading of prostate cancer tissue microarrays (TMAs)~\cite{arvaniti2018gleason}. The task involves multi-class classification of prostate tissue images into four categories: ``benign'', ``Gleason pattern 3'', ``Gleason pattern 4'', and ``Gleason pattern 5''. The dataset comprises 22,752 H\&E-stained images (750$\times$750 pixels at 40$\times$ magnification) from 886 patients across two cohorts: a training cohort of 641 patients and an independent test cohort of 245 patients, with annotations provided by two experienced pathologists. Following the \textit{eva}~\cite{kaiko.ai2024eva} protocol, the dataset is split into 15,303 training images, 2,482 validation images selected from TMA 76 due to its balanced distribution of Gleason scores, and 4,967 test images.

\paragraph{MoNuSAC} The MoNuSAC (Multi-Organ Nuclei Segmentation and Classification Challenge) dataset was introduced for the task of nuclear segmentation and classification across multiple organs and tissue types~\cite{monusac}. The task involves segmenting and classifying cell nuclei into five categories: ``epithelial cells'', ``lymphocytes'', ``macrophages'', ``neutrophils'', and ``ambiguous'' (cells that cannot be definitively classified into the other categories). 
The dataset consists of 294 H\&E-stained tissue images from four organs: breast, kidney, lung, and prostate, with variable dimensions, from 113$\times$81 to 1,398$\times$1,956 pixels at 40$\times$ magnification. The images were collected from 37 hospitals and 71 patients, containing over 46,000 annotated nuclei. Instance-level segmentation masks are provided, indicating both the spatial extent and class label of each nucleus. The dataset is split into 209 training images and 85 test images. Following the \textit{eva}~\cite{kaiko.ai2024eva} evaluation protocol, we use the test split as the validation split.

\paragraph{MSI CRC (patch)} This dataset contains 173,630 H\&E images (224$\times$224 pixels at 20$\times$ magnification) extracted from $N =$ 360 colorectal cancer (CRC) tissue scans from TCGA (TCGA-CRC-DX). The task is binary classification of microsatellite instability (MSI) vs.\ microsatellite stability (MSS), which is a clinically important prognostic marker~\cite{kather2019deep,kaczmarzyk2023champkit}. The dataset is split into 56,044 (28,022 MSI + 28,022 MSS) training images, 18,682 (9,341 MSI + 9,341 MSS) validation images, and 98,904 (28,335 MSI + 70,569 MSS) test images.

\paragraph{MSI STAD (patch)} This dataset contains 198,464 H\&E images (224$\times$224 pixels at 20$\times$ magnification) extracted from $N =$ 315 stomach adenocarcinoma (STAD) tissue scans from TCGA (TCGA-STAD). The task is binary classification of microsatellite instability (MSI) vs.\ microsatellite stability (MSS), which is a clinically important prognostic marker~\cite{kather2019deep,kaczmarzyk2023champkit}. The dataset is split into 60,342 (30,171 MSI + 30,171 MSS) training images, 20,114 (10,057 MSI + 10,057 MSS) validation images, and 118,008 (27,904 MSI + 90,104 MSS) test images.

\paragraph{TCGA Uniform (10$\times$) and (20$\times$)} The TCGA Uniform dataset~\cite{komura_tcga_uniform} contains 264,110 to 271,700 patches per resolution with $256\times256$ pixels. The task is to differentiate between 32 cancer classes from different tissue types (e.g., Colon adenocarcinoma). Only patches showing the specific cancer type were extracted from the TCGA WSIs based on annotations from two trained pathologists.
As there is no official train and test split, we generate five folds with no overlapping patients and sample 100 patches per class and fold, resulting in a total dataset size of 16,000 patches. We generate one dataset containing patches with \SI{0.5}{\micro\metre}/pixel (20$\times$) and one with \SI{1.0}{\micro\metre}/pixel (10$\times$) to test the performance of the foundation models at different zoom levels. The results represent the mean balanced accuracy for the five-fold cross-validation evaluation. 

\paragraph{PathoROB} The PathoROB Robustness Index was introduced by~\cite{koemen2025pathorob} as a novel metric to quantify the robustness of pathology foundation models to non-biological confounding features, specifically medical center differences arising from variations in staining procedures, scanner hardware, surgical techniques, and laboratory protocols. The Robustness Index measures the degree to which biological features (e.g., tissue type, cancer type) dominate over confounding non-biological features (e.g., medical center signatures) in the neighborhood structure of a foundation model's embedding space. The benchmark comprises three datasets sourced from CAMELYON~\cite{camelyon,camelyon17}, TCGA~\cite{komura_tcga_uniform}, and Tolkach ESCA~\cite{tolkach2023esca}, covering 28 biological classes from 34 medical centers.

\paragraph{Plismbench} The Plismbench robustness benchmark was introduced by~\cite{plismbench} as a framework to quantify the robustness of pathology foundation models to technical variations arising from different scanning devices and staining procedures. The benchmark measures robustness through cosine similarity and top-k retrieval accuracy metrics, evaluating the ability of foundation models to maintain consistent representations of matching tissue tiles across different technical conditions. The evaluation is performed on registered tile pairs from the PLISM dataset~\cite{ochi2024registered}, which consists of 46 human tissue types processed with 13 different H\&E staining conditions and captured using 7 WSI scanners, resulting in 16,278 registered tissue tile pairs. Robustness is assessed across three conditions: cross-scanner robustness (fixed staining), cross-staining robustness (fixed scanner), and combined cross-scanner and cross-staining robustness.

\paragraph{Patho-Bench} The Patho-Bench framework was introduced by~\cite{pathobench} for comprehensive benchmarking of patch-level and slide-level foundation models for whole-slide images (WSIs) across diverse clinically relevant pathology tasks. The benchmark comprises 95 tasks, subcategorized into seven clinical prediction categories: morphological subtyping (11 tasks), tumor microenvironment (TME) characterization (16 tasks), tumor grading (9 tasks), molecular subtyping (6 tasks), mutation prediction (34 tasks), treatment response and assessment (7 tasks), and survival prediction (12 tasks). We evaluate on a subset of 53 tasks drawn from multiple datasets, including CPTAC datasets covering breast (BRCA), colorectal (COAD), glioblastoma (GBM), head and neck (HNSC), lung squamous and adenocarcinoma (LSCC, LUAD), pancreatic (PDA), clear cell renal cell carcinoma (ccRCC), BC-Therapy for breast cancer, MBC for metastatic breast cancer, Hancock for head and neck squamous cell carcinoma, IMP-Cervical for cervical cancer, and Boehmke for ovarian cancer.
Each task is formulated as a WSI-level classification or survival prediction problem. Following the Patho-Bench~\cite{pathobench} evaluation protocol, we use the Attention-based Multiple Instance Learning (ABMIL) framework~\cite{abmil}, where H\&E-stained whole-slide images are represented as bags of patch-level feature embeddings extracted from foundation models. We employ the pre-defined task folds, depending on cohort size with label- and patient-stratified splits.

\subsection{Additional results}
\label{ssec:additional_results}

In Table~\ref{tab:main_results_cls} and Table~\ref{tab:distilled_results_cls} we present results for the same table contents and formatting as in Table~\ref{tab:main_results} and Table~\ref{tab:distillation_results} but using the CLS token instead of CLS+MEAN. Moreover, in Table~\ref{tab:all_models_results_cls_mean} and Table~\ref{tab:all_models_results_cls} we directly compare all large-scale and all resource-efficient pathology foundation models for both CLS token and the CLS+MEAN embedding.

\begin{table}[h]

\centering
\begin{adjustbox}{width=1\textwidth}
\scriptsize
\begin{tabular}{l>{\raggedleft\arraybackslash}p{0.5cm}>{\raggedleft\arraybackslash}p{0.5cm}>{\raggedleft\arraybackslash}p{0.5cm}>{\raggedleft\arraybackslash}p{0.5cm}>{\raggedleft\arraybackslash}p{0.5cm}>{\raggedleft\arraybackslash}p{0.5cm}>{\raggedleft\arraybackslash}p{0.5cm}>{\raggedleft\arraybackslash}p{0.5cm}>{\raggedleft\arraybackslash}p{0.5cm}>{\raggedleft\arraybackslash}p{0.5cm}>{\raggedleft\arraybackslash}p{0.5cm}}%
&\rotatebox{50}{\makebox[0pt][l]{\hspace{-3mm}\strut \makecell[r]{Atlas 2}}}&\rotatebox{50}{\makebox[0pt][l]{\hspace{-3mm}\strut \makecell[r]{Pluto-4G*}}}&\rotatebox{50}{\makebox[0pt][l]{\hspace{-3mm}\strut \makecell[r]{Atlas}}}&\rotatebox{50}{\makebox[0pt][l]{\hspace{-3mm}\strut \makecell[r]{UNI2-H}}}&\rotatebox{50}{\makebox[0pt][l]{\hspace{-3mm}\strut \makecell[r]{H-Optimus-0}}}&\rotatebox{50}{\makebox[0pt][l]{\hspace{-3mm}\strut \makecell[r]{Midnight-12k}}}&\rotatebox{50}{\makebox[0pt][l]{\hspace{-3mm}\strut \makecell[r]{Virchow2}}}&\rotatebox{50}{\makebox[0pt][l]{\hspace{-3mm}\strut \makecell[r]{Prov-GigaPath}}}&\rotatebox{50}{\makebox[0pt][l]{\hspace{-3mm}\strut \makecell[r]{UNI}}}&\rotatebox{50}{\makebox[0pt][l]{\hspace{-3mm}\strut \makecell[r]{Phikon v2}}}&\rotatebox{50}{\makebox[0pt][l]{\hspace{-3mm}\strut \makecell[r]{H-Optimus-1*}}}\\%
\addlinespace[0.3cm]%
\toprule%
HEST{-}ccRCC&26.0&\textbf{28.9}&\underline{27.8}&26.4&26.8&21.3&27.4&24.7&24.0&26.6&24.5\\%
HEST{-}COAD&\textbf{32.3}&31.6&25.9&30.1&30.9&29.1&25.9&30.2&26.2&25.0&\underline{32.0}\\%
HEST{-}IDC&\textbf{62.1}&\underline{60.6}&59.6&59.0&59.8&58.2&59.2&56.5&57.4&54.1&60.2\\%
HEST{-}LUNG&\underline{57.2}&56.9&57.0&55.8&55.9&55.8&55.3&54.2&54.6&54.2&\textbf{57.8}\\%
HEST{-}LYMPH{-}IDC&\textbf{28.2}&27.3&25.7&27.2&25.9&26.4&25.6&25.1&25.6&24.4&\underline{27.7}\\%
HEST{-}PAAD&\textbf{52.3}&\underline{51.1}&50.7&50.0&49.1&49.0&47.2&48.6&48.1&44.5&49.6\\%
HEST{-}PRAD&\textbf{38.5}&37.4&35.3&35.7&\textbf{38.5}&33.7&34.8&37.0&29.4&35.5&\underline{37.8}\\%
HEST{-}RECTUM&\underline{24.0}&23.3&21.3&22.3&22.2&18.5&20.9&19.5&18.4&17.5&\textbf{24.2}\\%
HEST{-}SKCM&\textbf{67.8}&\underline{67.0}&56.2&65.9&64.5&63.6&61.9&57.6&63.5&55.5&65.9\\%
MSI CRC (patch)&\textbf{78.2}&{-}&71.7&\underline{71.8}&69.7&69.9&71.6&69.0&69.1&67.6&{-}\\%
MSI STAD (patch)&\textbf{75.8}&{-}&\underline{73.5}&72.8&72.6&72.6&73.0&67.5&68.6&68.6&{-}\\%
\addlinespace[0.3cm]%
HEST{-}Average&\textbf{43.2}&\underline{42.7}&39.9&41.4&41.5&39.5&39.8&39.3&38.6&37.5&42.2\\%
Molecular{-}Average&\textbf{49.3}&{-}&45.9&\underline{47.0}&46.9&45.3&45.7&44.5&44.1&43.0&{-}\\%
\addlinespace[0.3cm]%
\hline%
\addlinespace[0.3cm]%
BACH&91.1&\textbf{93.8}&\underline{93.0}&91.5&75.9&90.5&88.3&75.5&78.5&73.3&{-}\\%
BreakHis&\underline{85.6}&81.5&\textbf{91.1}&85.5&80.1&81.2&82.1&82.8&78.9&71.3&{-}\\%
CoNSeP&\textbf{66.1}&65.0&\underline{65.1}&63.6&64.5&62.4&64.5&63.0&63.0&62.9&{-}\\%
CRC{-}100k&\textbf{97.1}&96.4&\underline{96.9}&96.5&95.5&96.6&96.7&95.1&94.4&93.9&95.6\\%
Gleason&\underline{79.5}&79.3&78.7&77.5&77.0&\textbf{80.0}&78.3&72.4&75.0&75.7&{-}\\%
MHIST&\textbf{88.0}&\underline{87.5}&86.1&82.6&84.4&80.4&86.1&83.0&84.4&77.7&83.5\\%
MoNuSAC&\underline{69.4}&\textbf{70.4}&67.4&64.5&68.1&65.8&66.8&66.4&65.0&64.3&{-}\\%
PANDA&\textbf{68.3}&66.8&65.8&64.6&\underline{67.2}&65.2&64.9&65.5&65.1&62.4&{-}\\%
PCAM&\textbf{96.1}&95.1&\underline{95.3}&95.0&94.3&92.9&93.8&94.5&93.7&89.4&{-}\\%
CAMELYON16 (0.25 MPP)&\textbf{86.5}&{-}&85.5&85.2&83.1&84.0&\underline{85.9}&81.0&83.2&80.2&{-}\\%
TCGA Uniform (1.0 MPP)&\textbf{84.2}&{-}&78.7&79.3&75.6&\underline{82.7}&78.2&74.3&74.0&69.8&{-}\\%
TCGA Uniform (0.5 MPP)&\textbf{82.6}&{-}&76.8&78.8&78.7&\underline{82.5}&77.8&74.0&74.5&76.7&{-}\\%
\addlinespace[0.3cm]%
Morphology{-}Average{-}Pluto{-}4G&\textbf{82.4}&81.8&\underline{82.2}&80.1&78.6&79.4&80.2&77.6&77.6&74.5&{-}\\%
Morphology{-}Average&\textbf{82.9}&{-}&\underline{81.7}&80.4&78.7&80.4&80.3&77.3&77.5&74.8&{-}\\%
\addlinespace[0.3cm]%
\hline%
\addlinespace[0.3cm]%
Prediction{-}Average{-}Pluto{-}4G&\textbf{62.8}&\underline{62.2}&61.0&60.8&60.0&59.5&60.0&58.4&58.1&56.0&{-}\\%
Prediction{-}Average&\textbf{66.8}&{-}&\underline{64.6}&64.4&63.5&63.6&63.7&61.6&61.5&59.6&{-}\\%
\addlinespace[0.3cm]%
\hline%
\addlinespace[0.3cm]%
\addlinespace[0.3cm]%
\addlinespace[0.3cm]%
\hline%
\addlinespace[0.3cm]%
Robustness Index (Camelyon)&\textbf{93.6}&{-}&\underline{80.0}&51.5&69.0&46.7&74.1&36.0&10.8&1.6&{-}\\%
Robustness Index (TCGA)&\textbf{87.8}&{-}&82.3&79.6&80.4&\underline{85.4}&82.7&72.7&72.9&58.8&{-}\\%
Robustness Index (Tolkach)&\textbf{95.9}&{-}&93.2&91.4&91.0&93.9&\underline{95.2}&68.9&87.6&73.3&{-}\\%
Plismbench (Leaderboard metric)&\textbf{60.5}&{-}&35.4&33.2&\underline{48.0}&33.8&44.8&33.3&32.5&16.4&{-}\\%
\addlinespace[0.3cm]%
Robustness{-}Average&\textbf{84.4}&{-}&72.7&63.9&72.1&65.0&\underline{74.2}&52.7&51.0&37.5&{-}\\%
\addlinespace[0.3cm]%
\hline%
\addlinespace[0.1cm]%
\multicolumn{12}{r}{*Results for Pluto-4G taken from~\cite{pluto4g}. Results for H-Optimus-1 taken from \cite{hoptimus1}.}
\end{tabular}
\end{adjustbox}
\vspace{3mm}
\caption{Same table content and formatting as in Table~\ref{tab:main_results}, but for the CLS token instead of CLS+MEAN.}
\label{tab:main_results_cls}
\end{table}

\begin{table}[h]
\centering
\begin{adjustbox}{width=1.0\textwidth}
\scriptsize
\begin{tabular}{l>{\raggedleft\arraybackslash}p{0.5cm}@{\hspace{5mm}\vrule}>{\raggedleft\arraybackslash}p{0.5cm}>{\raggedleft\arraybackslash}p{0.5cm}>{\raggedleft\arraybackslash}p{0.5cm}>{\raggedleft\arraybackslash}p{0.5cm}>{\raggedleft\arraybackslash}p{0.5cm}@{\hspace{5mm}\vrule}>{\raggedleft\arraybackslash}p{0.5cm}>{\raggedleft\arraybackslash}p{0.5cm}>{\raggedleft\arraybackslash}p{0.5cm}>{\raggedleft\arraybackslash}p{0.5cm}>{\raggedleft\arraybackslash}p{0.5cm}}%
&&\multicolumn{5}{c|}{\textbf{ViT-B}}&\multicolumn{5}{c}{\textbf{ViT-S}}\\[0.8cm]%
&\rotatebox{50}{\makebox[0pt][l]{\hspace{-3mm}\strut \makecell[r]{Atlas 2}}}&\rotatebox{50}{\makebox[0pt][l]{\hspace{-3mm}\strut \makecell[r]{Atlas 2-B}}}&\rotatebox{50}{\makebox[0pt][l]{\hspace{-3mm}\strut \makecell[r]{ViT-B/8 \cite{kaiko2024towardslarge}}}}&\rotatebox{50}{\makebox[0pt][l]{\hspace{-3mm}\strut \makecell[r]{Hibou-B}}}&\rotatebox{50}{\makebox[0pt][l]{\hspace{-3mm}\strut \makecell[r]{Phikon}}}&\rotatebox{50}{\makebox[0pt][l]{\hspace{-3mm}\strut \makecell[r]{H0-mini*}}}&\rotatebox{50}{\makebox[0pt][l]{\hspace{-3mm}\strut \makecell[r]{Atlas 2-S}}}&\rotatebox{50}{\makebox[0pt][l]{\hspace{-3mm}\strut \makecell[r]{Pluto-4S-8*}}}&\rotatebox{50}{\makebox[0pt][l]{\hspace{-3mm}\strut \makecell[r]{Pluto-4S-16*}}}&\rotatebox{50}{\makebox[0pt][l]{\hspace{-3mm}\strut \makecell[r]{ViT-S/8 \cite{kang2022benchmarking}}}}&\rotatebox{50}{\makebox[0pt][l]{\hspace{-3mm}\strut \makecell[r]{ViT-S/8 \cite{kaiko2024towardslarge}}}}\\[8pt]%
\toprule%
HEST{-}ccRCC&26.0&\underline{26.5}&23.1&20.6&24.2&\textbf{26.7}&\textbf{27.6}&{-}&{-}&\underline{25.5}&23.5\\%
HEST{-}COAD&32.3&\textbf{30.4}&\underline{26.8}&25.9&26.2&24.9&\textbf{30.8}&{-}&{-}&\underline{27.6}&22.8\\%
HEST{-}IDC&62.1&\textbf{60.4}&56.0&54.8&53.3&\underline{58.6}&\textbf{59.8}&{-}&{-}&\underline{54.8}&53.0\\%
HEST{-}LUNG&57.2&\textbf{55.1}&51.8&54.1&54.7&\underline{54.8}&\underline{54.0}&{-}&{-}&\textbf{54.1}&50.5\\%
HEST{-}LYMPH{-}IDC&28.2&\textbf{27.0}&22.7&23.7&23.7&\underline{26.3}&\textbf{26.2}&{-}&{-}&\underline{25.4}&24.6\\%
HEST{-}PAAD&52.3&\textbf{50.1}&46.0&46.1&44.2&\underline{49.2}&\textbf{47.3}&{-}&{-}&\underline{42.2}&41.8\\%
HEST{-}PRAD&38.5&\underline{36.1}&\underline{36.1}&29.7&34.2&\textbf{36.8}&\textbf{35.2}&{-}&{-}&25.0&\underline{33.4}\\%
HEST{-}RECTUM&24.0&\textbf{22.8}&16.2&16.4&15.3&\underline{18.6}&\textbf{19.9}&{-}&{-}&\underline{14.9}&14.7\\%
HEST{-}SKCM&67.8&\textbf{62.6}&57.3&54.4&53.6&\underline{60.1}&51.7&{-}&{-}&\textbf{58.1}&\underline{51.8}\\%
MSI CRC (patch)&78.2&\textbf{75.2}&\underline{68.6}&68.0&67.4&{-}&\textbf{74.2}&{-}&{-}&\underline{71.0}&68.4\\%
MSI STAD (patch)&75.8&\textbf{76.1}&68.7&\underline{70.2}&67.3&{-}&\textbf{77.4}&{-}&{-}&\underline{72.4}&70.4\\%
\addlinespace[0.3cm]%
HEST{-}Average&43.2&\textbf{41.2}&37.3&36.2&36.6&\underline{39.6}&\textbf{39.2}&\underline{36.5}&36.2&36.4&35.1\\%
Molecular{-}Average&49.3&\textbf{47.5}&\underline{43.0}&42.2&42.2&{-}&\textbf{45.8}&{-}&{-}&\underline{42.8}&41.4\\%
\addlinespace[0.3cm]%
\hline%
\addlinespace[0.3cm]%
BACH&91.1&\textbf{90.4}&\underline{86.6}&82.5&73.0&77.4&\textbf{90.6}&\underline{82.7}&79.6&76.3&81.7\\%
BreakHis&85.6&\textbf{90.8}&\underline{83.2}&74.1&70.8&{-}&\textbf{84.2}&\underline{81.3}&79.2&68.8&71.7\\%
CoNSeP&66.1&\textbf{65.8}&\underline{64.5}&63.3&62.8&62.9&\textbf{65.9}&\underline{64.9}&62.1&63.4&63.9\\%
CRC{-}100k&97.1&\textbf{96.8}&95.6&95.1&94.0&\underline{96.1}&\textbf{96.9}&\underline{95.2}&95.0&94.5&\underline{95.2}\\%
Gleason&79.5&\textbf{78.3}&\underline{74.4}&72.9&72.9&{-}&\textbf{79.1}&76.2&\underline{76.3}&74.4&71.7\\%
MHIST&88.0&\textbf{86.6}&\underline{82.9}&79.0&80.3&79.0&\underline{83.6}&\textbf{83.7}&83.5&74.8&81.8\\%
MoNuSAC&69.4&\textbf{69.3}&\underline{68.5}&64.9&64.3&64.3&\underline{67.2}&\textbf{67.8}&64.0&66.7&65.3\\%
PANDA&68.3&\textbf{67.7}&63.4&62.4&64.5&\underline{66.7}&\textbf{66.6}&61.8&61.5&\underline{61.9}&60.4\\%
PCAM&96.1&\textbf{95.6}&91.5&\underline{94.5}&92.0&94.2&\textbf{94.5}&90.7&90.9&\underline{92.0}&88.6\\%
CAMELYON16 (0.25 MPP)&86.5&\textbf{87.2}&81.8&77.1&81.2&\underline{84.2}&\textbf{85.9}&{-}&{-}&78.4&\underline{80.3}\\%
TCGA Uniform (1.0 MPP)&84.2&\textbf{81.2}&\underline{80.3}&72.2&70.5&{-}&\textbf{76.8}&{-}&{-}&71.6&\underline{73.8}\\%
TCGA Uniform (0.5 MPP)&82.6&\textbf{79.4}&\underline{79.3}&70.5&75.5&{-}&\textbf{75.6}&{-}&{-}&71.3&\underline{71.6}\\%
\addlinespace[0.3cm]%
Morphology{-}Average{-}Pluto{-}4G&82.4&\textbf{82.4}&\underline{79.0}&76.5&75.0&{-}&\textbf{81.0}&\underline{78.3}&76.9&74.8&75.6\\%
Morphology{-}Average&82.9&\textbf{82.4}&\underline{79.3}&75.7&75.2&{-}&\textbf{80.6}&{-}&{-}&74.5&\underline{75.5}\\%
\addlinespace[0.3cm]%
\hline%
\addlinespace[0.3cm]%
Prediction{-}Average{-}Pluto{-}4G&62.8&\textbf{61.8}&\underline{58.1}&56.4&55.8&{-}&\textbf{60.1}&\underline{57.4}&56.6&55.6&55.4\\%
Prediction{-}Average&66.8&\textbf{65.7}&\underline{62.0}&59.7&59.4&{-}&\textbf{64.0}&{-}&{-}&\underline{59.4}&59.2\\%
\addlinespace[0.3cm]%
\hline%
\addlinespace[0.3cm]%
\addlinespace[0.3cm]%
\addlinespace[0.3cm]%
\hline%
\addlinespace[0.3cm]%
Robustness Index (Camelyon)&93.6&\textbf{95.0}&\underline{13.2}&4.8&0.9&{-}&\textbf{95.5}&{-}&{-}&3.6&\underline{29.6}\\%
Robustness Index (TCGA)&87.8&\textbf{88.0}&\underline{75.9}&59.7&60.0&{-}&\textbf{87.4}&{-}&{-}&65.8&\underline{74.6}\\%
Robustness Index (Tolkach)&95.9&\textbf{95.3}&\underline{88.8}&75.5&76.1&{-}&\textbf{95.9}&{-}&{-}&82.3&\underline{92.2}\\%
Plismbench (Leaderboard metric)&60.5&\textbf{58.5}&32.6&20.4&19.3&\underline{54.1}&\textbf{58.2}&{-}&{-}&30.0&\underline{36.3}\\%
\addlinespace[0.3cm]%
Robustness{-}Average&84.4&\textbf{84.2}&\underline{52.6}&40.1&39.1&{-}&\textbf{84.2}&{-}&{-}&45.4&\underline{58.2}\\%
\addlinespace[0.3cm]%
\hline%
\addlinespace[0.1cm]%
\multicolumn{12}{r}{*Results for H0-mini taken from~\cite{plismbench} (\textit{eva}), \cite{plismbenchgithub} (Plismbench) and \cite{hestgithub} (HEST). Results for Pluto-4S-8 and Pluto-4S-16 taken from \cite{pluto4g}.}
\end{tabular}%

\end{adjustbox}
\vspace{3mm}
\caption{Same table content and formatting as in Table~\ref{tab:distillation_results}, but for the CLS token instead of CLS+MEAN.}
\label{tab:distilled_results_cls}
\end{table}

\begin{table}[p]
\centering
\rotatebox{90}{%
\begin{minipage}{\textheight}
\centering
\scriptsize
\resizebox{\textwidth}{!}{%
\begin{tabular}{l>{\raggedleft\arraybackslash}p{0.5cm}>{\raggedleft\arraybackslash}p{0.5cm}>{\raggedleft\arraybackslash}p{0.5cm}>{\raggedleft\arraybackslash}p{0.5cm}>{\raggedleft\arraybackslash}p{0.5cm}>{\raggedleft\arraybackslash}p{0.5cm}>{\raggedleft\arraybackslash}p{0.5cm}>{\raggedleft\arraybackslash}p{0.5cm}>{\raggedleft\arraybackslash}p{0.5cm}>{\raggedleft\arraybackslash}p{0.5cm}>{\raggedleft\arraybackslash}p{0.5cm}>{\raggedleft\arraybackslash}p{0.5cm}>{\raggedleft\arraybackslash}p{0.5cm}>{\raggedleft\arraybackslash}p{0.5cm}>{\raggedleft\arraybackslash}p{0.5cm}>{\raggedleft\arraybackslash}p{0.5cm}>{\raggedleft\arraybackslash}p{0.5cm}>{\raggedleft\arraybackslash}p{0.5cm}>{\raggedleft\arraybackslash}p{0.5cm}>{\raggedleft\arraybackslash}p{0.5cm}>{\raggedleft\arraybackslash}p{0.5cm}>{\raggedleft\arraybackslash}p{0.5cm}>{\raggedleft\arraybackslash}p{0.5cm}}%
&\rotatebox{50}{\makebox[0pt][l]{\hspace{-3mm}\strut \makecell[r]{Atlas 2}}}&\rotatebox{50}{\makebox[0pt][l]{\hspace{-3mm}\strut \makecell[r]{Atlas 2-B}}}&\rotatebox{50}{\makebox[0pt][l]{\hspace{-3mm}\strut \makecell[r]{Atlas 2-S}}}&\rotatebox{50}{\makebox[0pt][l]{\hspace{-3mm}\strut \makecell[r]{Pluto-4G*}}}&\rotatebox{50}{\makebox[0pt][l]{\hspace{-3mm}\strut \makecell[r]{Atlas}}}&\rotatebox{50}{\makebox[0pt][l]{\hspace{-3mm}\strut \makecell[r]{UNI2-H}}}&\rotatebox{50}{\makebox[0pt][l]{\hspace{-3mm}\strut \makecell[r]{H-Optimus-0}}}&\rotatebox{50}{\makebox[0pt][l]{\hspace{-3mm}\strut \makecell[r]{Midnight-12k}}}&\rotatebox{50}{\makebox[0pt][l]{\hspace{-3mm}\strut \makecell[r]{Virchow2}}}&\rotatebox{50}{\makebox[0pt][l]{\hspace{-3mm}\strut \makecell[r]{Prov-GigaPath}}}&\rotatebox{50}{\makebox[0pt][l]{\hspace{-3mm}\strut \makecell[r]{ViT-B/8 \cite{kaiko2024towardslarge}}}}&\rotatebox{50}{\makebox[0pt][l]{\hspace{-3mm}\strut \makecell[r]{UNI}}}&\rotatebox{50}{\makebox[0pt][l]{\hspace{-3mm}\strut \makecell[r]{Pluto-4S-8*}}}&\rotatebox{50}{\makebox[0pt][l]{\hspace{-3mm}\strut \makecell[r]{Hibou-B}}}&\rotatebox{50}{\makebox[0pt][l]{\hspace{-3mm}\strut \makecell[r]{ViT-B/16 \cite{kaiko2024towardslarge}}}}&\rotatebox{50}{\makebox[0pt][l]{\hspace{-3mm}\strut \makecell[r]{Phikon-v2}}}&\rotatebox{50}{\makebox[0pt][l]{\hspace{-3mm}\strut \makecell[r]{Pluto-4S-16*}}}&\rotatebox{50}{\makebox[0pt][l]{\hspace{-3mm}\strut \makecell[r]{Phikon}}}&\rotatebox{50}{\makebox[0pt][l]{\hspace{-3mm}\strut \makecell[r]{ViT-S/8 \cite{kang2022benchmarking}}}}&\rotatebox{50}{\makebox[0pt][l]{\hspace{-3mm}\strut \makecell[r]{ViT-S/8 \cite{kaiko2024towardslarge}}}}&\rotatebox{50}{\makebox[0pt][l]{\hspace{-3mm}\strut \makecell[r]{ViT-S/16 \cite{kaiko2024towardslarge}}}}&\rotatebox{50}{\makebox[0pt][l]{\hspace{-3mm}\strut \makecell[r]{ViT-S/16 \cite{kang2022benchmarking}}}}&\rotatebox{50}{\makebox[0pt][l]{\hspace{-3mm}\strut \makecell[r]{H0-mini*}}}\\%
\addlinespace[0.3cm]%
\toprule%
HEST{-}ccRCC&27.9&26.8&26.4&{-}&\textbf{29.5}&28.1&\underline{29.2}&20.9&27.2&24.9&24.8&25.1&{-}&22.8&23.4&27.4&{-}&24.2&26.7&25.7&23.7&26.1&26.4\\%
HEST{-}COAD&\textbf{34.9}&32.4&32.0&{-}&29.3&\underline{33.2}&30.9&31.8&25.8&32.0&30.2&25.8&{-}&28.7&28.0&25.6&{-}&27.8&30.2&23.0&24.3&27.6&27.0\\%
HEST{-}IDC&\textbf{62.7}&\underline{61.2}&60.8&{-}&60.4&60.5&61.1&59.9&59.7&58.1&57.0&58.5&{-}&56.0&55.7&56.8&{-}&54.8&55.2&54.5&54.0&53.1&59.1\\%
HEST{-}LUNG&\textbf{59.4}&56.8&54.9&{-}&58.1&57.4&57.5&\underline{58.3}&56.8&56.1&53.9&55.6&{-}&54.9&53.5&55.0&{-}&56.7&54.4&53.2&53.7&52.2&56.3\\%
HEST{-}LYMPH{-}IDC&\textbf{28.6}&27.1&26.6&{-}&26.2&27.4&26.6&\underline{27.5}&25.7&25.0&23.5&25.8&{-}&25.0&23.7&24.8&{-}&23.8&24.9&23.9&23.0&25.6&26.4\\%
HEST{-}PAAD&\textbf{54.2}&52.0&49.7&{-}&51.7&\underline{52.3}&51.1&50.2&47.8&49.9&47.8&48.8&{-}&45.9&46.2&47.7&{-}&46.5&44.2&43.3&44.7&44.4&50.7\\%
HEST{-}PRAD&\textbf{39.7}&38.8&35.7&{-}&38.6&37.5&36.2&37.1&35.3&\underline{39.1}&38.0&32.7&{-}&30.6&32.5&37.9&{-}&34.5&27.1&34.4&33.4&27.6&36.3\\%
HEST{-}RECTUM&\textbf{25.4}&\underline{24.5}&20.8&{-}&22.6&22.7&24.0&20.3&20.7&19.4&17.0&17.6&{-}&17.6&15.5&18.7&{-}&16.6&15.5&14.4&14.5&13.8&20.5\\%
HEST{-}SKCM&\textbf{70.0}&63.4&60.8&{-}&62.0&\underline{68.3}&66.1&64.8&64.0&59.7&59.4&62.3&{-}&55.9&55.5&58.4&{-}&54.5&57.5&56.5&53.9&54.9&61.2\\%
MSI CRC (patch)&\textbf{78.0}&\underline{74.9}&\underline{74.9}&{-}&72.6&72.0&70.0&70.8&71.9&69.5&68.8&69.7&{-}&68.3&68.0&68.3&{-}&67.2&70.2&69.4&68.1&69.6&{-}\\%
MSI STAD (patch)&\underline{76.1}&75.0&\textbf{78.1}&{-}&73.0&73.1&72.2&73.7&72.5&68.1&68.1&68.6&{-}&71.0&68.9&69.2&{-}&67.1&72.6&72.3&70.1&68.7&{-}\\%
\addlinespace[0.3cm]%
HEST{-}Average&\textbf{44.8}&42.6&40.9&\underline{43.2}&42.0&43.0&42.5&41.2&40.3&40.5&39.1&39.1&36.9&37.5&37.1&39.1&36.4&37.7&37.3&36.5&36.1&36.1&40.4\\%
Molecular{-}Average&\textbf{50.6}&\underline{48.4}&47.3&{-}&47.6&\underline{48.4}&47.7&46.8&46.1&45.6&44.4&44.6&{-}&43.3&42.8&44.5&{-}&43.1&43.5&42.8&42.1&42.1&{-}\\%
\addlinespace[0.3cm]%
\hline%
\addlinespace[0.3cm]%
BACH&91.7&90.9&90.4&\textbf{93.2}&\underline{92.3}&92.2&74.7&90.6&88.6&79.2&88.0&79.7&85.1&81.9&83.0&73.6&79.8&73.5&77.3&80.2&84.1&77.7&{-}\\%
BreakHis&88.3&\textbf{90.9}&84.0&81.8&\underline{89.9}&86.3&81.0&83.4&81.1&84.5&84.4&80.4&80.8&79.8&82.5&70.9&76.8&71.0&72.7&73.0&76.5&75.7&{-}\\%
CoNSeP&\textbf{66.1}&65.8&\underline{65.9}&65.0&65.1&63.6&64.5&62.4&64.5&63.0&64.5&63.0&64.9&63.3&60.6&62.9&62.1&62.8&63.4&63.9&60.3&60.0&62.9\\%
CRC{-}100k&\textbf{97.2}&97.0&\underline{97.1}&96.8&96.9&96.9&96.4&96.7&96.6&95.9&96.0&96.1&95.4&95.8&95.9&95.3&95.5&94.7&94.9&95.1&94.5&94.0&{-}\\%
Gleason&\textbf{80.7}&78.7&\underline{80.5}&78.5&78.4&77.6&77.3&79.0&78.5&72.7&75.4&76.0&76.3&73.6&74.1&75.3&76.6&74.2&74.3&73.3&70.8&75.9&{-}\\%
MHIST&\textbf{88.5}&\underline{88.0}&84.7&87.9&86.8&82.6&85.1&81.2&86.5&83.3&83.0&83.8&84.2&79.9&84.3&79.6&83.4&82.3&76.5&82.7&82.8&78.6&{-}\\%
MoNuSAC&\underline{69.4}&69.3&67.2&\textbf{70.4}&67.4&64.5&68.1&65.8&66.8&66.4&68.5&65.0&67.8&64.9&63.5&64.3&64.0&64.3&66.7&65.3&63.3&62.1&64.3\\%
PANDA&\underline{68.3}&\textbf{68.4}&67.2&66.6&66.4&66.2&66.2&65.4&64.7&65.0&64.7&65.4&63.1&63.2&62.5&62.1&63.0&64.5&61.9&60.3&62.4&60.4&{-}\\%
PCAM&\textbf{96.1}&\underline{95.6}&94.5&95.2&95.2&95.1&94.4&93.0&93.9&94.6&92.0&93.9&91.4&94.3&90.9&90.0&91.6&92.8&92.4&88.6&90.5&90.2&{-}\\%
CAMELYON16 (0.25 MPP)&\underline{86.8}&\textbf{87.5}&86.0&{-}&86.3&85.9&84.7&86.1&85.7&80.7&81.9&84.1&{-}&80.3&80.8&80.7&{-}&82.8&79.1&81.1&80.0&77.4&{-}\\%
TCGA Uniform (1.0 MPP)&\textbf{84.1}&81.4&77.0&{-}&78.7&79.4&76.0&\underline{83.7}&78.0&74.7&80.5&74.3&{-}&72.4&78.1&70.1&{-}&70.8&71.9&74.0&75.3&68.8&{-}\\%
TCGA Uniform (0.5 MPP)&\underline{82.7}&79.5&75.6&{-}&77.0&78.8&78.8&\textbf{83.2}&77.5&74.3&79.2&74.5&{-}&71.1&77.5&76.7&{-}&75.6&71.5&72.2&75.4&67.9&{-}\\%
\addlinespace[0.3cm]%
Morphology{-}Average{-}Pluto{-}4G&\textbf{82.9}&\underline{82.7}&81.3&81.7&82.0&80.6&78.6&79.7&80.1&78.3&79.6&78.1&78.8&77.4&77.5&74.9&77.0&75.6&75.6&75.8&76.1&75.0&{-}\\%
Morphology{-}Average&\textbf{83.3}&\underline{82.8}&80.8&{-}&81.7&80.8&78.9&80.9&80.2&77.9&79.8&78.0&{-}&76.7&77.8&75.1&{-}&75.8&75.2&75.8&76.3&74.1&{-}\\%
\addlinespace[0.3cm]%
\hline%
\addlinespace[0.3cm]%
Prediction{-}Average{-}Pluto{-}4G&\textbf{63.8}&\underline{62.6}&61.1&62.5&62.0&61.8&60.6&60.5&60.2&59.4&59.3&58.6&57.8&57.4&57.3&57.0&56.7&56.6&56.4&56.2&56.1&55.6&{-}\\%
Prediction{-}Average&\textbf{67.7}&\underline{66.3}&64.8&{-}&65.4&65.3&64.0&64.6&63.9&62.4&62.9&62.0&{-}&60.7&61.1&60.5&{-}&60.1&60.0&60.0&60.0&58.8&{-}\\%
\addlinespace[0.3cm]%
\hline%
\addlinespace[0.3cm]%
\addlinespace[0.3cm]%
\addlinespace[0.3cm]%
\hline%
\addlinespace[0.3cm]%
Robustness Index (Camelyon)&94.0&\underline{95.3}&\textbf{95.6}&{-}&78.5&54.4&70.5&47.8&79.9&38.2&14.7&14.5&{-}&5.2&9.7&1.9&{-}&1.1&4.3&26.5&8.9&12.0&71.8\\%
Robustness Index (TCGA)&\underline{87.9}&\textbf{88.2}&87.5&{-}&82.6&80.3&81.2&85.8&82.2&73.7&76.3&74.7&{-}&61.3&71.2&61.9&{-}&62.3&66.1&74.0&66.0&68.7&79.4\\%
Robustness Index (Tolkach)&\textbf{96.4}&95.9&\underline{96.3}&{-}&93.8&92.3&91.8&94.1&95.4&74.6&89.6&90.2&{-}&79.5&87.6&76.8&{-}&79.5&83.2&91.8&87.0&87.5&93.2\\%
Plismbench (Leaderboard metric)&\underline{64.5}&\textbf{65.3}&64.1&{-}&41.1&36.0&52.2&37.3&46.4&37.3&35.8&38.5&{-}&24.6&31.2&19.6&{-}&24.4&32.1&37.4&33.0&40.0&{-}\\%
\addlinespace[0.3cm]%
Robustness{-}Average&85.7&\textbf{86.2}&\underline{85.9}&{-}&74.0&65.8&73.9&66.2&76.0&56.0&54.1&54.5&{-}&42.6&49.9&40.0&{-}&41.8&46.4&57.4&48.7&52.0&{-}\\%
\addlinespace[0.3cm]%
\hline%
\addlinespace[0.1cm]%
\multicolumn{24}{r}{*Results for H0-mini taken from~\cite{plismbench} (HEST, \textit{eva}) and \cite{pathorobgithub} (PathoROB). Results for Pluto-4G, Pluto-4S-8 and Pluto-4S-16 taken from~\cite{pluto4g}.}
\end{tabular}%

}
\vspace{3mm}
\caption{Prediction results for tasks with morphology and molecular targets as well as for robustness benchmarks. The table shows results for all models from Table~\ref{tab:main_results}\&\ref{tab:distillation_results}, as well as for additional resource efficient models that use a patch-token size of 16. The \textbf{best} result per row is in bold, the \underline{second-best} result is underlined.}
\label{tab:all_models_results_cls_mean}
\end{minipage}
}
\end{table}

\begin{table}[p]
\centering
\rotatebox{90}{%
\begin{minipage}{\textheight}
\centering
\scriptsize
\resizebox{\textwidth}{!}{%
\begin{tabular}{l>{\raggedleft\arraybackslash}p{0.5cm}>{\raggedleft\arraybackslash}p{0.5cm}>{\raggedleft\arraybackslash}p{0.5cm}>{\raggedleft\arraybackslash}p{0.5cm}>{\raggedleft\arraybackslash}p{0.5cm}>{\raggedleft\arraybackslash}p{0.5cm}>{\raggedleft\arraybackslash}p{0.5cm}>{\raggedleft\arraybackslash}p{0.5cm}>{\raggedleft\arraybackslash}p{0.5cm}>{\raggedleft\arraybackslash}p{0.5cm}>{\raggedleft\arraybackslash}p{0.5cm}>{\raggedleft\arraybackslash}p{0.5cm}>{\raggedleft\arraybackslash}p{0.5cm}>{\raggedleft\arraybackslash}p{0.5cm}>{\raggedleft\arraybackslash}p{0.5cm}>{\raggedleft\arraybackslash}p{0.5cm}>{\raggedleft\arraybackslash}p{0.5cm}>{\raggedleft\arraybackslash}p{0.5cm}>{\raggedleft\arraybackslash}p{0.5cm}>{\raggedleft\arraybackslash}p{0.5cm}>{\raggedleft\arraybackslash}p{0.5cm}>{\raggedleft\arraybackslash}p{0.5cm}>{\raggedleft\arraybackslash}p{0.5cm}>{\raggedleft\arraybackslash}p{0.5cm}}%
&\rotatebox{50}{\makebox[0pt][l]{\hspace{-3mm}\strut \makecell[r]{Atlas 2}}}&\rotatebox{50}{\makebox[0pt][l]{\hspace{-3mm}\strut \makecell[r]{Atlas 2-B}}}&\rotatebox{50}{\makebox[0pt][l]{\hspace{-3mm}\strut \makecell[r]{Atlas 2-S}}}&\rotatebox{50}{\makebox[0pt][l]{\hspace{-3mm}\strut \makecell[r]{Pluto-4G*}}}&\rotatebox{50}{\makebox[0pt][l]{\hspace{-3mm}\strut \makecell[r]{Atlas}}}&\rotatebox{50}{\makebox[0pt][l]{\hspace{-3mm}\strut \makecell[r]{UNI2-H}}}&\rotatebox{50}{\makebox[0pt][l]{\hspace{-3mm}\strut \makecell[r]{H-Optimus-0}}}&\rotatebox{50}{\makebox[0pt][l]{\hspace{-3mm}\strut \makecell[r]{Midnight-12k}}}&\rotatebox{50}{\makebox[0pt][l]{\hspace{-3mm}\strut \makecell[r]{Virchow2}}}&\rotatebox{50}{\makebox[0pt][l]{\hspace{-3mm}\strut \makecell[r]{Prov-GigaPath}}}&\rotatebox{50}{\makebox[0pt][l]{\hspace{-3mm}\strut \makecell[r]{ViT-B/8 \cite{kaiko2024towardslarge}}}}&\rotatebox{50}{\makebox[0pt][l]{\hspace{-3mm}\strut \makecell[r]{UNI}}}&\rotatebox{50}{\makebox[0pt][l]{\hspace{-3mm}\strut \makecell[r]{Pluto-4S-8*}}}&\rotatebox{50}{\makebox[0pt][l]{\hspace{-3mm}\strut \makecell[r]{Hibou-B}}}&\rotatebox{50}{\makebox[0pt][l]{\hspace{-3mm}\strut \makecell[r]{ViT-B/16 \cite{kaiko2024towardslarge}}}}&\rotatebox{50}{\makebox[0pt][l]{\hspace{-3mm}\strut \makecell[r]{Phikon-v2}}}&\rotatebox{50}{\makebox[0pt][l]{\hspace{-3mm}\strut \makecell[r]{Pluto-4S-16*}}}&\rotatebox{50}{\makebox[0pt][l]{\hspace{-3mm}\strut \makecell[r]{Phikon}}}&\rotatebox{50}{\makebox[0pt][l]{\hspace{-3mm}\strut \makecell[r]{ViT-S/8 \cite{kang2022benchmarking}}}}&\rotatebox{50}{\makebox[0pt][l]{\hspace{-3mm}\strut \makecell[r]{ViT-S/8 \cite{kaiko2024towardslarge}}}}&\rotatebox{50}{\makebox[0pt][l]{\hspace{-3mm}\strut \makecell[r]{ViT-S/16 \cite{kaiko2024towardslarge}}}}&\rotatebox{50}{\makebox[0pt][l]{\hspace{-3mm}\strut \makecell[r]{ViT-S/16 \cite{kang2022benchmarking}}}}&\rotatebox{50}{\makebox[0pt][l]{\hspace{-3mm}\strut \makecell[r]{H-Optimus-1*}}}&\rotatebox{50}{\makebox[0pt][l]{\hspace{-3mm}\strut \makecell[r]{H0-mini*}}}\\%
\addlinespace[0.3cm]%
\toprule%
HEST{-}ccRCC&26.0&26.5&27.6&\textbf{28.9}&\underline{27.8}&26.4&26.8&21.3&27.4&24.7&23.1&24.0&{-}&20.6&22.9&26.6&{-}&24.2&25.5&23.5&21.0&24.3&24.5&26.7\\%
HEST{-}COAD&\textbf{32.3}&30.4&30.8&31.6&25.9&30.1&30.9&29.1&25.9&30.2&26.8&26.2&{-}&25.9&28.1&25.0&{-}&26.2&27.6&22.8&20.6&25.3&\underline{32.0}&24.9\\%
HEST{-}IDC&\textbf{62.1}&60.4&59.8&\underline{60.6}&59.6&59.0&59.8&58.2&59.2&56.5&56.0&57.4&{-}&54.8&53.5&54.1&{-}&53.3&54.8&53.0&53.3&52.4&60.2&58.6\\%
HEST{-}LUNG&\underline{57.2}&55.1&54.0&56.9&57.0&55.8&55.9&55.8&55.3&54.2&51.8&54.6&{-}&54.1&51.6&54.2&{-}&54.7&54.1&50.5&50.3&53.0&\textbf{57.8}&54.8\\%
HEST{-}LYMPH{-}IDC&\textbf{28.2}&27.0&26.2&27.3&25.7&27.2&25.9&26.4&25.6&25.1&22.7&25.6&{-}&23.7&23.7&24.4&{-}&23.7&25.4&24.6&22.5&24.8&\underline{27.7}&26.3\\%
HEST{-}PAAD&\textbf{52.3}&50.1&47.3&\underline{51.1}&50.7&50.0&49.1&49.0&47.2&48.6&46.0&48.1&{-}&46.1&45.2&44.5&{-}&44.2&42.2&41.8&44.1&43.4&49.6&49.2\\%
HEST{-}PRAD&\textbf{38.5}&36.1&35.2&37.4&35.3&35.7&\textbf{38.5}&33.7&34.8&37.0&36.1&29.4&{-}&29.7&32.8&35.5&{-}&34.2&25.0&33.4&34.8&28.2&\underline{37.8}&36.8\\%
HEST{-}RECTUM&\underline{24.0}&22.8&19.9&23.3&21.3&22.3&22.2&18.5&20.9&19.5&16.2&18.4&{-}&16.4&15.3&17.5&{-}&15.3&14.9&14.7&13.3&13.7&\textbf{24.2}&18.6\\%
HEST{-}SKCM&\textbf{67.8}&62.6&51.7&\underline{67.0}&56.2&65.9&64.5&63.6&61.9&57.6&57.3&63.5&{-}&54.4&55.0&55.5&{-}&53.6&58.1&51.8&54.5&52.6&65.9&60.1\\%
MSI CRC (patch)&\textbf{78.2}&\underline{75.2}&74.2&{-}&71.7&71.8&69.7&69.9&71.6&69.0&68.6&69.1&{-}&68.0&68.6&67.6&{-}&67.4&71.0&68.4&68.1&70.0&{-}&{-}\\%
MSI STAD (patch)&75.8&\underline{76.1}&\textbf{77.4}&{-}&73.5&72.8&72.6&72.6&73.0&67.5&68.7&68.6&{-}&70.2&70.4&68.6&{-}&67.3&72.4&70.4&69.5&69.1&{-}&{-}\\%
\addlinespace[0.3cm]%
HEST{-}Average&\textbf{43.2}&41.2&39.2&\underline{42.7}&39.9&41.4&41.5&39.5&39.8&39.3&37.3&38.6&36.5&36.2&36.5&37.5&36.2&36.6&36.4&35.1&34.9&35.3&42.2&39.6\\%
Molecular{-}Average&\textbf{49.3}&\underline{47.5}&45.8&{-}&45.9&47.0&46.9&45.3&45.7&44.5&43.0&44.1&{-}&42.2&42.5&43.0&{-}&42.2&42.8&41.4&41.1&41.5&{-}&{-}\\%
\addlinespace[0.3cm]%
\hline%
\addlinespace[0.3cm]%
BACH&91.1&90.4&90.6&\textbf{93.8}&\underline{93.0}&91.5&75.9&90.5&88.3&75.5&86.6&78.5&82.7&82.5&83.6&73.3&79.6&73.0&76.3&81.7&81.2&77.0&{-}&77.4\\%
BreakHis&85.6&\underline{90.8}&84.2&81.5&\textbf{91.1}&85.5&80.1&81.2&82.1&82.8&83.2&78.9&81.3&74.1&80.0&71.3&79.2&70.8&68.8&71.7&73.0&73.9&{-}&{-}\\%
CoNSeP&\textbf{66.1}&65.8&\underline{65.9}&65.0&65.1&63.6&64.5&62.4&64.5&63.0&64.5&63.0&64.9&63.3&60.6&62.9&62.1&62.8&63.4&63.9&60.3&60.0&{-}&62.9\\%
CRC{-}100k&\textbf{97.1}&96.8&\underline{96.9}&96.4&\underline{96.9}&96.5&95.5&96.6&96.7&95.1&95.6&94.4&95.2&95.1&95.7&93.9&95.0&94.0&94.5&95.2&94.1&94.0&95.6&96.1\\%
Gleason&\underline{79.5}&78.3&79.1&79.3&78.7&77.5&77.0&\textbf{80.0}&78.3&72.4&74.4&75.0&76.2&72.9&73.0&75.7&76.3&72.9&74.4&71.7&69.8&75.0&{-}&{-}\\%
MHIST&\textbf{88.0}&86.6&83.6&\underline{87.5}&86.1&82.6&84.4&80.4&86.1&83.0&82.9&84.4&83.7&79.0&83.7&77.7&83.5&80.3&74.8&81.8&83.0&78.1&83.5&79.0\\%
MoNuSAC&\underline{69.4}&69.3&67.2&\textbf{70.4}&67.4&64.5&68.1&65.8&66.8&66.4&68.5&65.0&67.8&64.9&63.5&64.3&64.0&64.3&66.7&65.3&63.3&62.1&{-}&64.3\\%
PANDA&\textbf{68.3}&\underline{67.7}&66.6&66.8&65.8&64.6&67.2&65.2&64.9&65.5&63.4&65.1&61.8&62.4&62.5&62.4&61.5&64.5&61.9&60.4&61.9&61.0&{-}&66.7\\%
PCAM&\textbf{96.1}&\underline{95.6}&94.5&95.1&95.3&95.0&94.3&92.9&93.8&94.5&91.5&93.7&90.7&94.5&90.6&89.4&90.9&92.0&92.0&88.6&90.2&89.8&{-}&94.2\\%
CAMELYON16 (0.25 MPP)&\underline{86.5}&\textbf{87.2}&85.9&{-}&85.5&85.2&83.1&84.0&85.9&81.0&81.8&83.2&{-}&77.1&81.4&80.2&{-}&81.2&78.4&80.3&78.6&75.9&{-}&84.2\\%
TCGA Uniform (1.0 MPP)&\textbf{84.2}&81.2&76.8&{-}&78.7&79.3&75.6&\underline{82.7}&78.2&74.3&80.3&74.0&{-}&72.2&78.0&69.8&{-}&70.5&71.6&73.8&75.3&68.6&{-}&{-}\\%
TCGA Uniform (0.5 MPP)&\textbf{82.6}&79.4&75.6&{-}&76.8&78.8&78.7&\underline{82.5}&77.8&74.0&79.3&74.5&{-}&70.5&77.4&76.7&{-}&75.5&71.3&71.6&75.2&68.0&{-}&{-}\\%
\addlinespace[0.3cm]%
Morphology{-}Average{-}Pluto{-}4G&\textbf{82.4}&\textbf{82.4}&81.0&81.8&\underline{82.2}&80.1&78.6&79.4&80.2&77.6&79.0&77.6&78.3&76.5&77.0&74.5&76.9&75.0&74.8&75.6&75.2&74.5&{-}&{-}\\%
Morphology{-}Average&\textbf{82.9}&\underline{82.4}&80.6&{-}&81.7&80.4&78.7&80.4&80.3&77.3&79.3&77.5&{-}&75.7&77.5&74.8&{-}&75.2&74.5&75.5&75.5&73.6&{-}&{-}\\%
\addlinespace[0.3cm]%
\hline%
\addlinespace[0.3cm]%
Prediction{-}Average{-}Pluto{-}4G&\textbf{62.8}&61.8&60.1&\underline{62.2}&61.0&60.8&60.0&59.5&60.0&58.4&58.1&58.1&57.4&56.4&56.7&56.0&56.6&55.8&55.6&55.4&55.1&54.9&{-}&{-}\\%
Prediction{-}Average&\textbf{66.8}&\underline{65.7}&64.0&{-}&64.6&64.4&63.5&63.6&63.7&61.6&62.0&61.5&{-}&59.7&60.7&59.6&{-}&59.4&59.4&59.2&59.0&58.3&{-}&{-}\\%
\addlinespace[0.3cm]%
\hline%
\addlinespace[0.3cm]%
\addlinespace[0.3cm]%
\addlinespace[0.3cm]%
\hline%
\addlinespace[0.3cm]%
Robustness Index (Camelyon)&93.6&\underline{95.0}&\textbf{95.5}&{-}&80.0&51.5&69.0&46.7&74.1&36.0&13.2&10.8&{-}&4.8&9.2&1.6&{-}&0.9&3.6&29.6&8.7&10.3&{-}&{-}\\%
Robustness Index (TCGA)&\underline{87.8}&\textbf{88.0}&87.4&{-}&82.3&79.6&80.4&85.4&82.7&72.7&75.9&72.9&{-}&59.7&71.1&58.8&{-}&60.0&65.8&74.6&65.8&68.7&{-}&{-}\\%
Robustness Index (Tolkach)&\textbf{95.9}&\underline{95.3}&\textbf{95.9}&{-}&93.2&91.4&91.0&93.9&95.2&68.9&88.8&87.6&{-}&75.5&87.1&73.3&{-}&76.1&82.3&92.2&86.3&87.3&{-}&{-}\\%
Plismbench (Leaderboard metric)&\textbf{60.5}&\underline{58.5}&58.2&{-}&35.4&33.2&48.0&33.8&44.8&33.3&32.6&32.5&{-}&20.4&27.8&16.4&{-}&19.3&30.0&36.3&30.9&38.4&{-}&54.1\\%
\addlinespace[0.3cm]%
Robustness{-}Average&\textbf{84.4}&\underline{84.2}&\underline{84.2}&{-}&72.7&63.9&72.1&65.0&74.2&52.7&52.6&51.0&{-}&40.1&48.8&37.5&{-}&39.1&45.4&58.2&47.9&51.2&{-}&{-}\\%
\addlinespace[0.3cm]%
\hline%
\addlinespace[0.1cm]%
\multicolumn{25}{r}{*Results for Pluto-4G, Pluto-4S-8 and Pluto-4S-16 taken from \cite{pluto4g}. Results for H-Optimus-1 taken from~\cite{hoptimus1}. Results for H0-mini taken from~\cite{plismbench} (\textit{eva}), \cite{plismbenchgithub} (Plismbench) and \cite{hestgithub} (HEST).}
\end{tabular}%

}
\vspace{3mm}
\caption{Same table content and formatting as in Table~\ref{tab:all_models_results_cls_mean}, but for the CLS token instead of CLS+MEAN.}
\label{tab:all_models_results_cls}
\end{minipage}
}
\end{table}

\end{document}